\documentclass[11pt, a4paper, twocolumn, logo, nonumbering]{antgroup}
\usepackage[square, numbers]{natbib}
\usepackage{dblfloatfix}
\usepackage{caption}
\usepackage{graphicx}
\usepackage{float}
\usepackage{placeins}
\usepackage{xspace}
\usepackage{pifont}
\usepackage{multirow}

\usepackage{longtable}
\usepackage{hyperref}
\hypersetup{
  hidelinks,
  pdftitle={APTER: Adaptive Post-Training with Expert-Grounded Rubrics},
  pdfauthor={Xukai Wang, Liangqi Li, Zhiyue Xu, Jingang Zhou, Xiaoyu Shi, Jiansheng Cai, Bo Zhang, Zhe Li, Xu-Yao Zhang}
}
\usepackage{amsfonts}
\usepackage{amsmath}
\usepackage{amssymb}
\usepackage{adjustbox}

\usepackage[bottom]{footmisc}

\usepackage{dsfont}
\usepackage{array}
\usepackage{tabularx}
\usepackage[table]{xcolor}
\usepackage{booktabs}
\usepackage{listings}

\usepackage[most,skins,theorems]{tcolorbox}

\definecolor{customgray}{RGB}{230,230,230}
\definecolor{aptergreen}{HTML}{16803A}
\definecolor{ablationblue}{RGB}{232,242,252}

\newcommand{\apter}{APTER\xspace}
\newcommand{\gain}[1]{\textcolor{aptergreen}{\,(+#1)}}
\newcommand{\scoregain}[2]{\textbf{#1}\gain{#2}}

\floatstyle{ruled}
\newfloat{algorithm}{tbp}{loa}
\floatname{algorithm}{Algorithm}

\tcbset{
  aibox/.style={
    width=\linewidth,
    top=8pt,
    bottom=4pt,
    colback=gray!10!white,
    colframe=gray!50!black,
    colbacktitle=gray!70!black,
    enhanced,
    breakable,
    center,
    attach boxed title to top left={yshift=-0.1in,xshift=0.15in},
    boxed title style={boxrule=0pt,colframe=white,},
  }
}
\newtcolorbox{AIbox}[2][]{aibox,title=#2,#1}

\makeatletter
\def\@BTrule[#1]{%
  \ifx\longtable\undefined
    \let\@BTswitch\@BTnormal
  \else\ifx\hline\LT@hline
    \nobreak
    \let\@BTswitch\@BLTrule
  \else
     \let\@BTswitch\@BTnormal
  \fi\fi
  \global\@thisrulewidth=#1\relax
  \ifnum\@thisruleclass=\tw@\vskip\@aboverulesep\else
  \ifnum\@lastruleclass=\z@\vskip\@aboverulesep\else
  \ifnum\@lastruleclass=\@ne\vskip\doublerulesep\fi\fi\fi
  \@BTswitch}
\makeatother

\addto\extrasenglish{
}

 {\begin{list}{}%
         {\setlength{\leftmargin}{#1}}%
         \item[]%
 }
 {\end{list}}
\renewcommand{\today}{}

\title{\centering APTER: Adaptive Post-Training with Expert-Grounded Rubrics}

\author{
Xukai Wang$^{*}$,
Liangqi Li$^{*}$,
Zhiyue Xu,
Jingang Zhou,
Xiaoyu Shi,
Jiansheng Cai,\newline
Bo Zhang$^{\dag}$,
Zhe Li,
Xu-Yao Zhang$^{\dag}$
\\
\vspace{-6pt}
Ant Digital Technologies, Ant Group
\vspace{-32pt}
}

\renewcommand{\phi}{\varphi}

\renewcommand{\geq}{\geqslant}

\renewcommand{\epsilon}{\varepsilon}
\renewcommand{\imath}{\mathrm{i}}

\newlength{\restsubwidth}
\newlength{\restsubheight}
\newlength{\restsubmoreheight}
\newcommand{\rest}[2]{%
        \settowidth{\restsubwidth}{\ensuremath{#2}}
        \settoheight{\restsubheight}{\ensuremath{{}_{#2}}}
        \ensuremath{{#1\hskip 0.5pt}_{\vrule\kern2pt\parbox[b][%
        4pt][b]{\the\restsubwidth}{%
                        \ensuremath{{}_{#2}}}}}
        }

\begin{abstract}
As large language models enter professional domains, they must satisfy domain constraints, include critical evidence, and provide complete reasoning rather than merely produce fluent responses.
Existing post-training methods often rely on holistic preferences or outcome-level verification, while recent rubric-based methods usually generate rubrics independently for each query.
In specialized domains, such unconstrained rubrics may omit critical requirements and vary across samples, hindering the diagnosis and targeted repair of persistent capability deficiencies.
We propose APTER (Adaptive Post-Training with Expert-Grounded Rubrics), a framework that integrates structured domain knowledge into fine-grained evaluation, optimization, and diagnosis for specialized complex reasoning.
First, expert-grounded rubric construction starts from an expert criteria framework built by domain experts, where each criterion represents a stable professional capability.
For each query, APTER selects relevant criteria and instantiates them into query-level rubrics linked to their source criteria, turning reusable expert criteria into executable query-level supervision without reference answers.
Second, adaptive post-training uses rubric verdicts as both optimization and criterion-level diagnostic signals.
Aggregating low-scoring verdicts by criterion ID reveals persistent deficiencies and triggers targeted supervised fine-tuning updates during reinforcement learning.
Experiments on mathematical reasoning and medical question answering show consistent gains across both domains.
Across three model generations, APTER improves the mathematics and medical averages over the corresponding base models by up to $15.86$ and $8.04$ points, respectively.
Code and rubric datasets are available at \url{https://github.com/AntDT-APTER/APTER.git}.
\end{abstract}

\begin{document}
\maketitle
\footnotetext[1]{\fontsize{7.5}{9}\selectfont Equal contribution.}
\footnotetext[2]{\fontsize{7.5}{9}\selectfont Corresponding to boyuan.zb@antgroup.com and xyz@nlpr.ia.ac.cn.}
\enlargethispage{3\baselineskip}

\section{Introduction}

\begin{figure}[!t]
\centering
\includegraphics[width=1\columnwidth]{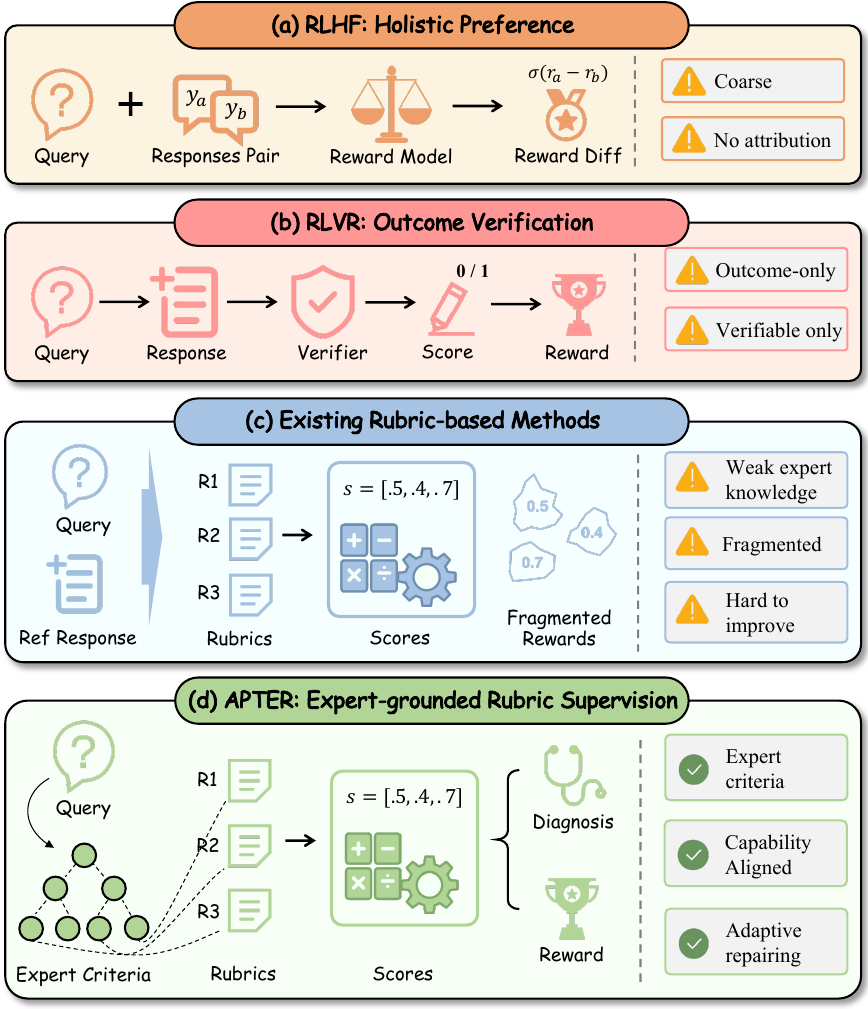}
\caption{Comparison of post-training supervision signals.
RLHF provides coarse response-level preferences and RLVR provides outcome verification, while existing rubric-based methods often generate query-level rubrics without explicit grounding in expert-defined criteria.
APTER integrates structured domain knowledge into rubric construction and uses rubric verdicts for fine-grained optimization, capability diagnosis, and targeted repair.}
\label{fig:apter-overview}
\end{figure}

Large language models (LLMs) increasingly address specialized tasks requiring
expert judgment~\cite{he2024olympiadbench,arora2025healthbench}.
In such settings, a fluent response with a plausible conclusion may still fail if it violates domain constraints, omits critical evidence, or lacks complete reasoning.
Recent analyses of medical LLM benchmarks further show that evaluating such failures requires lifecycle-oriented, safety-aware, and clinically faithful criteria, which are often specified by domain experts rather than captured by leaderboard-style final scores alone~\cite{ma2025beyond}.
As illustrated in Figure~\ref{fig:apter-overview}(a) and (b), existing post-training methods often use reinforcement learning from human feedback (RLHF) with pairwise preferences~\cite{ouyang2022training} or reinforcement learning with verifiable rewards (RLVR) for tasks with verifiable answers~\cite{cobbe2021training,shao2024deepseekmath}.
Both have proven effective, but expert knowledge is still insufficiently encoded in these signals: they do not explicitly represent which domain criteria should be checked, nor how failures on those criteria should guide model improvement.

Fine-grained supervision can better capture such professional requirements than holistic or outcome-level signals.
Process-level feedback offers one option~\cite{uesato2022solving,lightman2023lets,wang2023mathshepherd}, but it often requires expensive step-level annotations.
Rubric-based evaluation and LLM-as-a-Judge methods provide a more scalable alternative~\cite{zheng2023judging,gu2024survey,liu2025openrubrics,shen2026rethinking,gunjal2025rubrics,huang2025rubric}.
As shown in Figure~\ref{fig:apter-overview}(c), existing rubric-based methods often generate rubrics independently for individual queries or responses.
In specialized domains, however, rubrics should reflect domain-critical requirements that may be difficult for general-purpose LLMs to identify and prioritize from a query alone.
Unconstrained generation may omit critical constraints or encode inappropriate standards.
Variations in rubric granularity and domain depth also hinder aggregating failures across samples.
For example, missed contraindications and omitted safety constraints may appear unrelated unless mapped to a shared expert criterion such as safety-constraint handling.

We propose APTER (Adaptive Post-Training with Expert-Grounded Rubrics), a framework that integrates structured domain knowledge into fine-grained evaluation, optimization, and capability diagnosis for specialized complex reasoning.
APTER consists of expert-grounded rubric construction and adaptive post-training.
As illustrated in Figure~\ref{fig:apter-overview}(d), the first component uses an expert-constructed framework of reusable professional criteria, such as evidence coverage, constraint handling, and reasoning completeness.
It routes relevant criteria for each query and instantiates them as assessable rubrics with source criterion IDs, providing query-specific supervision without per-query expert authoring or reference answers.
The second uses rubric verdicts as both optimization and criterion-level diagnostic signals.
APTER supports Rubric RL, Rubric-based SFT, and Ada-IFT (Adaptive Interleaved Fine-Tuning).
Ada-IFT aggregates failures by criterion to identify persistent capability deficiencies and trigger targeted supervised repair during reinforcement learning, serving as the default configuration in the main experiments.
We evaluate APTER in two complementary regimes: mathematics tests rubric rewards against strong RLVR baselines in a verifiable setting, whereas open-ended medical question answering naturally requires multi-criterion expert evaluation and optimization.
Relative to the corresponding base models, APTER improves the mathematics macro-average by $6.13$--$15.86$ points and the medical macro-average by $5.19$--$8.04$ points across three Qwen model generations~\cite{qwen2.5,qwen3,qwen3.5}, with gains of up to $25.82$ points on AIME~24 and $17.96$ points on HealthBench.

Our main contributions are as follows:
\begin{enumerate}
    \item We introduce an expert-grounded rubric construction framework that integrates structured domain knowledge into query-level rubric generation by routing expert-defined criteria to each query and instantiating them as executable rubrics without requiring reference answers.
    \item We propose adaptive post-training, where rubric verdicts serve as both optimization and criterion-level diagnostic signals, enabling Ada-IFT to identify persistent capability deficiencies and trigger targeted repair.
    \item We implement APTER for mathematical reasoning and medical question answering, construct expert criteria frameworks and rubric datasets for both domains, and demonstrate consistent improvements over strong post-training baselines.
\end{enumerate}

\section{Related Work}
\label{sec:related}

\paragraph{Rubric construction and evaluation criteria.}
With the growth of LLM-as-a-Judge~\cite{zheng2023judging,gu2024survey} and
rubric-based reward modeling, recent work studies scalable rubric generation,
refinement, retrieval, and adaptive design~\cite{liu2025openrubrics,shen2026rethinking,zhou2026autochecklist,dhole2026rubricrag,ding2026adarubric}.
Domain evaluation frameworks such as HealthBench further highlight the value of aligning evaluation criteria with expert judgment in professional settings~\cite{arora2025healthbench}.
These methods commonly generate rubrics for individual queries or responses,
which is flexible but can vary in granularity and domain depth across samples.
APTER instead instantiates query-level rubrics from reusable expert criteria and
retains their criterion linkage for cross-query diagnosis.

\begin{figure*}[!t]
\centering
\includegraphics[width=\textwidth]{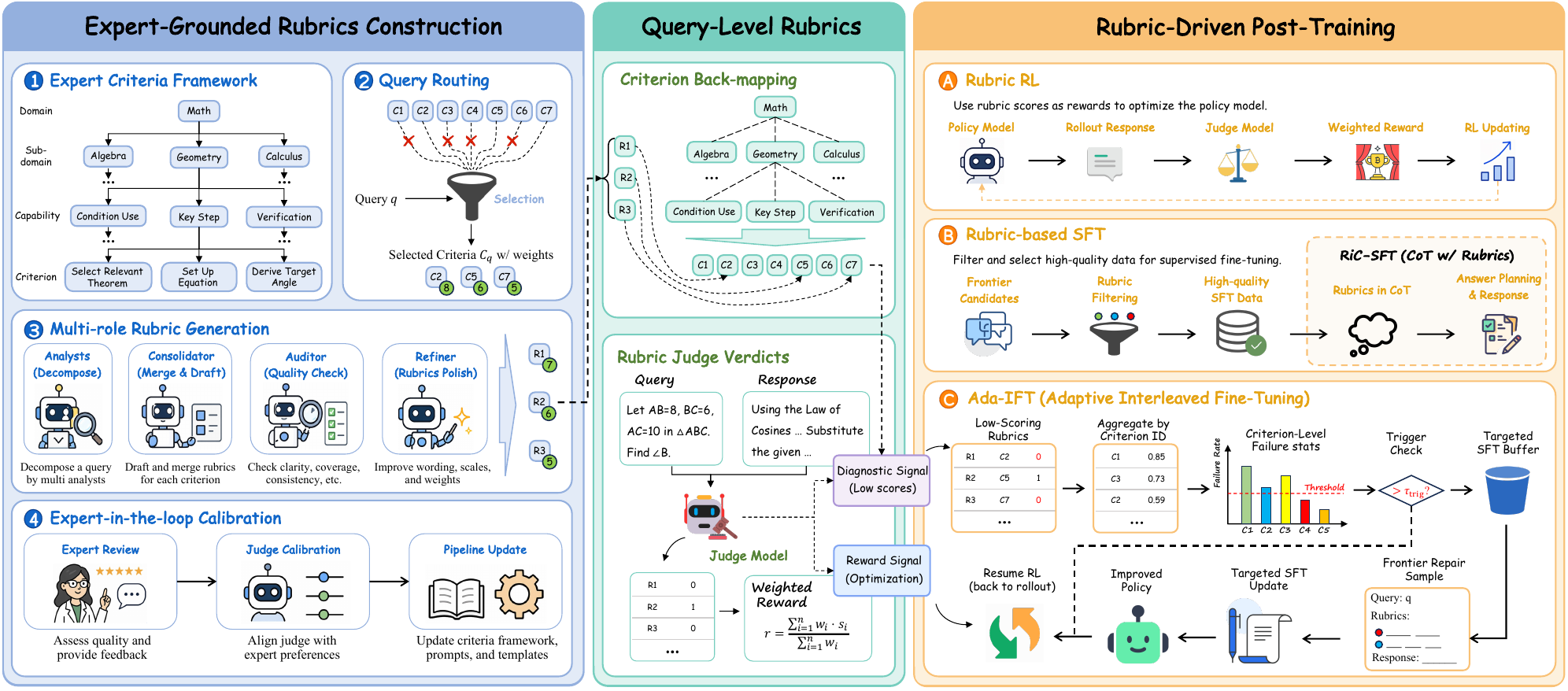}
\caption{
The APTER pipeline.
Expert-grounded rubric construction first builds an expert criteria framework and instantiates relevant criteria into query-level rubrics through routing, multi-role generation and refinement, and expert-in-the-loop calibration.
The resulting rubric verdicts preserve criterion back-mapping and are used for Rubric RL, Rubric-based SFT, and Ada-IFT.
}
\label{fig:method}
\end{figure*}

\paragraph{Rubric-based rewards for reasoning RL.}
Reasoning rewards range from outcome verification and
RLVR~\cite{cobbe2021training,shao2024deepseekmath,guo2025r1} to process-level
supervision~\cite{lightman2023lets,wang2023mathshepherd} and LLM- or
rubric-based evaluation for open-ended
responses~\cite{zheng2023judging,gunjal2025rubrics,huang2025rubric}.
Recent work incorporates rubric scores through reward design, rubric anchors,
or advantage decomposition~\cite{gunjal2025rubrics,huang2025rubric,yuan2025rrm,tan2026papo,lan2026arlrr}.
APTER likewise uses rubric-level rewards, while stable expert-criterion links
make the same verdicts reusable for capability diagnosis.

\paragraph{Interleaving RL with supervised fine-tuning.}
Interleaving RL with SFT can repair reasoning failures that exploration alone
does not resolve. ReST and rejection-sampling methods alternate sampling with
filtered SFT~\cite{gulcehre2023rest,yuan2023rft,xiong2025minimalist}, while
ReLIFT and iterative self-improvement introduce SFT during or between RL
stages~\cite{ma2025learning,deng2025openvlthinker}.
Rubric-guided and self-verification methods provide related signals for
near-miss responses~\cite{bi2025rgr,chen2026selfverify}; APTER instead
triggers repair from recurring failures aggregated under stable expert
criterion IDs.

\section{Methodology}
\label{sec:method}

\subsection{Overview}
\label{sec:method-overview}

We propose \apter (Adaptive Post-Training with Expert-Grounded Rubrics), a
framework combining expert-grounded rubric construction with adaptive
post-training. As shown in Figure~\ref{fig:method}, the former instantiates
expert-defined criteria into query-level rubrics with retained provenance; the
latter uses their verdicts for Rubric RL, Rubric-based SFT, and Ada-IFT, with
Ada-IFT serving as the default configuration.

\subsection{Expert-Grounded Rubric Construction}
\label{sec:rubrics-generation-optimization}

We call the resulting rubrics \emph{expert-grounded} because they are instantiated from an expert-constructed criteria framework, retain their source criterion IDs, and are calibrated through expert feedback.

\subsubsection{Expert Criteria Framework.}
\label{sec:expert-criteria-framework}

Before constructing query-level rubrics, domain experts organize recurring
professional capabilities into a hierarchy of domain $\rightarrow$ sub-domain
$\rightarrow$ capability category $\rightarrow$ criterion. Each reusable
criterion has a persistent identifier, a concise definition, and representative
examples. Experts review the framework for coverage, overlap, and operational
clarity, merging redundant criteria and clarifying ambiguous boundaries.

Expert-approved updates are incorporated between training runs, while the
framework and criterion IDs remain fixed within a run. The framework therefore
both constrains query-level rubric generation and provides a stable capability
space for aggregating verdicts across queries. The detailed construction,
review, and maintenance protocol is provided in the supplementary materials.

\subsubsection{Query-Level Routing and Rubric Instantiation.}
\label{sec:query-level-routing}

For each query, \apter selects relevant expert criteria and instantiates them into concrete, executable evaluation standards.
This process does not require experts to write every query-level rubric or provide reference answers, making it applicable to unlabeled and open-ended tasks without tying the evaluation standard to a single response.

Given a query $q$ and an expert criteria framework $C$, a routing function $P(\cdot)$ selects the subset of criteria relevant to the query:
\begin{equation}
C_q = P(q, C), \qquad C_q \subseteq C.
\end{equation}
For each selected criterion, \apter instantiates a query-specific rubric and
outputs a weighted rubric set:
\begin{equation}
r_i = G(q, c_i), \qquad
R_q = \{(c_i, w_i, r_i)\}_{i=1}^{L_q},
\end{equation}
where $G(\cdot)$ conditions generation on the query and selected criterion,
$L_q$ is the number of rubrics, and $w_i$ denotes relevance and importance.
The router excludes irrelevant criteria and preserves each criterion ID for
later aggregation.

\subsubsection{Multi-Role Rubric Generation and Refinement.}
After routing, Analyst LMs identify query-specific requirements under the
selected expert criteria. As shown in Figure~\ref{fig:method}, the Consolidator
merges their proposals, the Auditor checks clarity, assessability, relevance,
and redundancy, and the Refiner updates wording, scoring scales, and weights.
This separation reduces dependence on a single generation pass, while all
stages preserve source criterion IDs and the expert-defined capability space.

\subsubsection{Expert-in-the-Loop Calibration.}
Experts inspect query, response, and rubric samples and analyze expert--judge disagreements.
Rather than treating every disagreement as a judge error, experts distinguish among genuine policy-response failures, incorrect judge decisions, flawed query-level rubrics, and defects in the source criteria.
The corresponding actions include correcting judge supervision, revising a rubric or its weight, and proposing additions, removals, or revisions in the criteria framework.
This structured feedback is reused to improve rubric generation and judge calibration in subsequent iterations.
The framework and its identifiers remain fixed within each policy-training run; Ada-IFT uses them for targeted improvement but does not modify them automatically.
The detailed annotation decision flow and representative calibration cases are provided in the supplementary material.

\subsection{Adaptive Post-Training}
\label{sec:rubric-driven-post-training}

Let $\pi_\theta$ denote the policy model being post-trained and $J_\phi$ the judge model.
Given a query $q$, its rubric set $R_q$, and a response $y$ generated by $\pi_\theta$, the judge produces rubric-level scores conditioned on $(q,y,R_q)$; human annotations can replace or calibrate these scores when available.
\apter supports three ways to use this supervision.
Rubric RL aggregates the scores into scalar rewards for updating $\pi_\theta$; Rubric-based SFT uses them to select or structure supervised targets; and Ada-IFT combines reward-based optimization with criterion-level capability diagnosis and targeted repair.
Ada-IFT is the most distinctive strategy and serves as the default configuration in the main experiments.

\subsubsection{Rubric RL.}
\label{sec:rubrics-rl}

Rubric RL aggregates expert-grounded rubric scores into fine-grained rewards, extending recent rubric-based reward methods for open-ended domains~\cite{gunjal2025rubrics,huang2025rubric} and RL methods such as GRPO for verifiable reasoning~\cite{shao2024deepseekmath}.
The retained criterion provenance makes the resulting verdicts reusable for diagnosis in Ada-IFT.

In the RL stage, for each query $q$, $\pi_\theta$ samples $N$ rollouts, denoted as $Y_q = \{y^{(n)}\}_{n=1}^{N}$.
For each rollout $y^{(n)}$ and rubric $(c_i,w_i,r_i)\in R_q$, the judge
model outputs:
\begin{equation}
s_i^{(n)} = J_\phi(q, y^{(n)}, r_i).
\end{equation}

The $N$ rollouts and $L_q$ rubrics together form a verdict matrix $\mathbf{s} \in \mathbb{R}^{N \times L_q}$.
\apter then aggregates the rubric-level scores of the same response across different rubrics into a scalar reward for RL:
\begin{equation}
\mathrm{Reward}(q, y^{(n)}) =
\frac{\sum_{i=1}^{L_q} w_i s_i^{(n)}}{\sum_{i=1}^{L_q} w_i}.
\label{eq:rubric-reward}
\end{equation}

Compared with RLHF or RLVR methods that rely on holistic rewards or outcome-level verification, rubric rewards provide finer-grained and more interpretable rubric-level supervision by scoring specific reasoning and domain-compliance criteria.

\subsubsection{Rubric-based SFT.}
\label{sec:rubric-based-sft}

Rubric-based SFT uses expert-grounded rubrics to select reliable supervised data, following rejection-sampling-style post-training~\cite{xiong2025minimalist}.
For each query, the current policy $\pi_\theta$ generates multiple candidate responses $Y_q$, and $J_\phi$ evaluates each candidate using $R_q$.
\apter ranks the candidates with the same weighted score aggregation as Rubric RL, retains high-scoring responses as \texttt{<Query, Response>} pairs, and uses them to update $\pi_\theta$ with standard supervised learning.
Optionally, Rubric-in-CoT SFT (RiC-SFT) also places query-level rubrics before
the final answer as lightweight reasoning guidance~\cite{wei2022chain}, making
the evaluation requirements visible in the supervised trajectory. The sample
format is provided in the supplementary materials.

\subsubsection{Adaptive Interleaved Fine-Tuning.}
\label{sec:ada-ift}

\begin{figure}[!t]
\centering
\includegraphics[width=\columnwidth]{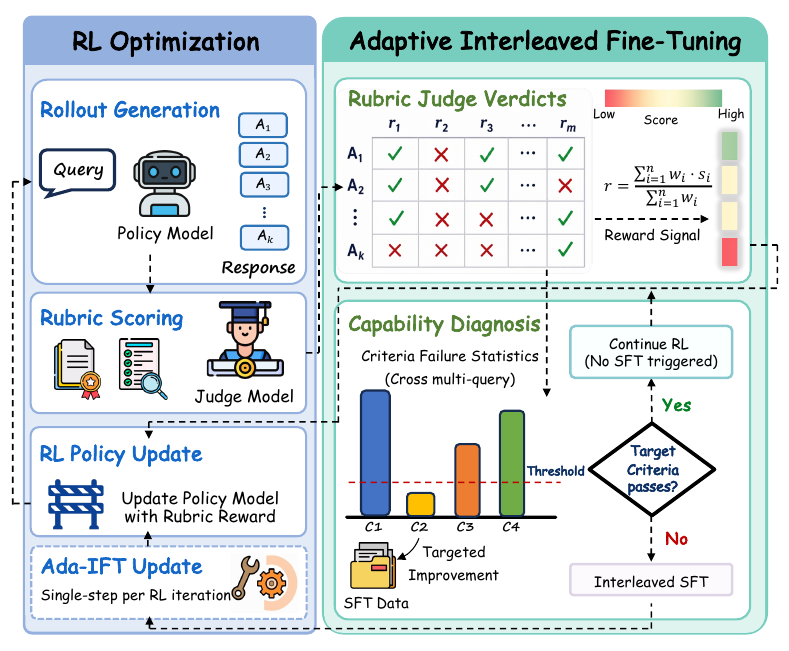}
\caption{Ada-IFT training flow. Rubric verdicts provide both a scalar reward
for the RL update and criterion-level diagnostic signals. Persistent
criterion failures trigger targeted SFT before training returns to RL.}
\label{fig:ada-ift}
\end{figure}

Ada-IFT extends Rubric RL with criterion-level capability diagnosis and dynamically triggered supervised repair.
Although Rubric RL provides fine-grained reward signals, RL exploration may be insufficient for persistent capability deficiencies.
Interleaved RL-SFT training has recently been explored for difficult reasoning questions and iterative reasoning improvement~\cite{ma2025learning,deng2025openvlthinker}.
\apter differs from these approaches by using expert-grounded criterion-level failure statistics as the trigger signal: targeted SFT is not activated merely by question difficulty or a fixed training curriculum, but by recurring failures on stable expert criteria.
Ada-IFT therefore uses the same judge verdicts both as scalar rewards for RL and as criterion-level diagnostic signals for targeted SFT.
Figure~\ref{fig:ada-ift} summarizes the two update channels acting on the same policy $\pi_\theta$: RL improves the policy continuously, while SFT is activated only when the diagnostic channel identifies a persistent criterion-level deficiency.

\begin{table*}[!t]
\centering
\small
\setlength{\tabcolsep}{1.4mm}
\begin{tabular*}{\textwidth}{@{\extracolsep{\fill}}llccccc@{}}
\toprule
\multicolumn{7}{c}{\textbf{(a) Mathematical Reasoning}} \\
\cmidrule(lr){1-7}
Model Family & Variant & AIME24 & AIME25 & MATH500 & Olympiad & Avg. \\
\midrule
Qwen2.5-7B-Instruct & Base
  & 13.33 & 4.61 & 68.84 & 38.71 & 31.37 \\
 & \apter
  & \scoregain{34.27}{20.94} & \scoregain{12.74}{8.13}
  & \scoregain{86.10}{17.26} & \scoregain{52.34}{13.63}
  & \scoregain{46.36}{14.99} \\
\addlinespace[2pt]
Qwen3-8B & Base
  & 23.33 & 18.33 & 84.26 & 55.89 & 45.45 \\
 & \apter
  & \scoregain{49.15}{25.82} & \scoregain{35.00}{16.67}
  & \scoregain{92.60}{8.34} & \scoregain{68.50}{12.61}
  & \scoregain{61.31}{15.86} \\
\addlinespace[2pt]
Qwen3.5-9B & Base
  & 62.00 & 53.13 & 97.20 & 78.64 & 72.74 \\
 & \apter
  & \scoregain{77.50}{15.50} & \scoregain{56.58}{3.45}
  & \scoregain{99.20}{2.00} & \scoregain{82.20}{3.56}
  & \scoregain{78.87}{6.13} \\
\bottomrule
\end{tabular*}

\begin{tabular*}{\textwidth}{@{\extracolsep{\fill}}llcccc@{}}
\toprule
\multicolumn{6}{c}{\textbf{(b) Medical Question Answering}} \\
\cmidrule(lr){1-6}
Model Family & Variant & HealthBench & MedQA & MedMCQA & Avg. \\
\midrule
Qwen2.5-7B-Instruct & Base
  & 31.95 & 58.99 & 54.82 & 48.59 \\
 & \apter
  & \scoregain{49.91}{17.96}
  & \scoregain{60.35}{1.36} & \scoregain{55.39}{0.57}
  & \scoregain{55.22}{6.63} \\
\addlinespace[2pt]
Qwen3-8B & Base
  & 44.90 & 63.94 & 58.64 & 55.83 \\
 & \apter
  & \scoregain{60.32}{15.42}
  & \scoregain{64.02}{0.08} & \scoregain{58.70}{0.06}
  & \scoregain{61.01}{5.19} \\
\addlinespace[2pt]
Qwen3.5-9B & Base
  & 54.26 & 76.42 & 68.01 & 66.23 \\
 & \apter
  & \scoregain{70.71}{16.45}
  & \scoregain{83.62}{7.20} & \scoregain{68.47}{0.46}
  & \scoregain{74.27}{8.04} \\
\bottomrule
\end{tabular*}
\caption{Main results across the Qwen2.5, Qwen3, and Qwen3.5 model
generations.  Each APTER row reports the final score followed by its absolute
improvement over the corresponding unmodified model in
\textcolor{aptergreen}{green parentheses}.  Mathematical reasoning and medical
question answering averages use four and three benchmarks, respectively.
All metrics are higher-is-better.}
\label{tab:main-opensource}
\end{table*}

\paragraph{Capability Diagnosis.}
The criterion ID attached to each rubric allows low-scoring verdicts to be aggregated across queries.
Ada-IFT uses two thresholds at different granularities.
Because the judge verdicts are binary, a rubric fails locally when
$s_i^{(n)}=0$. The step threshold $\tau_{\mathrm{step}}$ determines whether
the failure rate of a criterion within the current training step is high
enough to increment its persistent counter. The trigger threshold
$\tau_{\mathrm{trig}}$ specifies how many such step-level events must
accumulate before targeted repair.

For training step $t$, \apter first collects failed rubric instances:
\begin{equation}
\mathcal{E}^{(t)} =
\{(q, y^{(n)}, c_i, r_i) \mid s_i^{(n)}=0\}.
\end{equation}
For criterion $c$, let $\mathcal{O}_c^{(t)}$ denote all of its judged
occurrences in the step. Its step-level failure rate is
\begin{equation}
f_c^{(t)} =
\frac{\sum_{(q,y^{(n)},c_i,r_i)\in\mathcal{E}^{(t)}}
\mathbf{1}[c_i=c]}{|\mathcal{O}_c^{(t)}|}.
\end{equation}
Among criteria with $f_c^{(t)}\geq\tau_{\mathrm{step}}$, \apter increments
the counters of the top-$K$ failure rates. If a counter reaches
$\tau_{\mathrm{trig}}$, \apter treats that criterion as a persistent
capability deficiency.

\paragraph{Targeted SFT.}
Before policy training, a frontier model generates candidate repair responses
conditioned on the query, low-scoring rubric, source criterion, representative
policy failure, and diagnostic information; only verified responses are stored
in a criterion-indexed SFT database. When the counter of criterion $c$ reaches
$\tau_{\mathrm{trig}}$, \apter pauses RL, retrieves the corresponding
targeted samples $(q,y^*,c)$, and performs an SFT update. For medicine, the
retriever first attempts to use a verified response for the same query; for
mathematics, it samples criterion-matched records without replacement.
This update modifies $\pi_\theta$; training then returns to RL to test
persistence and discover other deficiencies. Hyperparameters are reported in
the supplementary material.


\section{Experiment}
\label{sec:experiment}

\newcommand{\placeablationtable}{%
\begin{table*}[!t]
\centering
\small
\setlength{\tabcolsep}{0.8mm}

\resizebox{\textwidth}{!}{%
\begin{tabular}{l l l l cccc}
\toprule
\multirow{2}{*}{Variant}
& \multicolumn{3}{c}{\textbf{Ablation Configuration}}
& \multicolumn{4}{c}{\textbf{Downstream Performance}} \\
\cmidrule(lr){2-4}\cmidrule(lr){5-8}
& Rubric Configuration & Reward Type & Training Strategy
& MATH500 & AIME24 & Olymp. & Health \\
\midrule
\rowcolor{ablationblue}
\multicolumn{8}{c}{\textbf{(A) Post-training Components}\quad
\textit{Rubric data, rubric reward, and Ada-IFT}} \\
Qwen3-8B
  & --- & --- & Base
  & 84.26 & 23.33 & 55.89 & 45.43 \\
SFT
  & Plain CoT data & --- & SFT
  & 84.30 & 26.79 & 55.71 & 44.62 \\
RiC-SFT
  & Rubric-augmented CoT data & --- & SFT
  & 88.86 & 28.37 & 57.05 & 47.10 \\
Rubric RL
  & Query-level rubric data & Rubric Reward & GRPO
  & 90.92 & 41.63 & 58.97 & 50.54 \\
\apter
  & Query-level rubric data & Rubric Reward & GRPO + Ada-IFT
  & \textbf{91.34} & \textbf{44.60} & \textbf{61.87} & \textbf{53.48} \\
\midrule
\rowcolor{ablationblue}
\multicolumn{8}{c}{\textbf{(B) Reward Type}\quad
\textit{Outcome-level ORM versus query-level rubric reward}} \\
GRPO
  & No rubrics & ORM Reward & GRPO
  & 89.16 & 37.75 & 58.65 & --- \\
Rubric RL
  & Query-level rubrics & Rubric Reward & GRPO
  & \textbf{90.92} & \textbf{41.63} & \textbf{58.97} & \textbf{50.54} \\
\midrule
\rowcolor{ablationblue}
\multicolumn{8}{c}{\textbf{(C) Rubric Granularity}\quad
\textit{No, generic, and query-level rubrics}} \\
GRPO
  & No rubrics & ORM Reward & GRPO
  & 89.16 & 37.75 & 58.65 & --- \\
Generic Criteria
  & Generic criteria & Rubric Reward & GRPO
  & 87.62 & 39.31 & 57.75 & 47.84 \\
Query-level Rubrics
  & Query-level rubrics & Rubric Reward & GRPO
  & \textbf{90.92} & \textbf{41.63} & \textbf{58.97} & \textbf{50.54} \\
\rowcolor{ablationblue}
\multicolumn{8}{c}{\textbf{(D) Rubric Provenance}\quad
\textit{Framework-free versus expert-grounded query-level rubrics}} \\
Framework-free
  & Framework-free rubrics & Rubric Reward & GRPO
  & 89.74 & 32.98 & 57.64 & 47.27 \\
Expert-grounded
  & Expert-grounded rubrics & Rubric Reward & GRPO
  & \textbf{90.92} & \textbf{41.63} & \textbf{58.97} & \textbf{50.54} \\
\midrule
\rowcolor{ablationblue}
\multicolumn{8}{c}{\textbf{(E) Rubric Refinement}\quad
\textit{Before versus after the Auditor--Refiner stage}} \\
Before Refine
  & Raw rubrics & Rubric Reward & GRPO
  & 86.92 & 36.67 & 56.10 & 48.18 \\
After Refine
  & Refined rubrics & Rubric Reward & GRPO
  & \textbf{90.92} & \textbf{41.63} & \textbf{58.97} & \textbf{50.54} \\
\bottomrule
\end{tabular}%
}
\caption{Unified summary of the five performance ablations on Qwen3-8B.
Light-blue section rows identify the factor isolated in each block.
The best result among fully reported comparable variants within each block is
highlighted in bold.}
\label{tab:ablation-unified}
\end{table*}
}

\subsection{Experimental Setup}
\label{sec:experimental-setup}

\paragraph{Models.}
We evaluate \apter on Qwen2.5-7B-Instruct~\cite{qwen2.5},
Qwen3-8B~\cite{qwen3}, and Qwen3.5-9B~\cite{qwen3.5}; all ablations use
Qwen3-8B. We use full-parameter fine-tuning in the \emph{non-thinking}
setting and keep prompts and output formats fixed across methods.

\paragraph{Training data and domains.}
For mathematics, we use $17$K problems from DAPO-Math~\cite{yu2025dapo}.
The main medical training set contains $14{,}713$ queries from
LiveMedBench~\cite{yan2026livemedbench},
SpeechMedDataset~\cite{chen2026speechmedassist}, and II-Medical
queries~\cite{intelligentinternet2025iimedical} distributed through
RubricHub~\cite{li2026rubrichub}. We use only the queries from RubricHub and
construct their rubrics with APTER. Main medical runs train for up to two
epochs and select checkpoints on HealthBench-500, a fixed random subset of
$500$ HealthBench examples~\cite{arora2025healthbench};
Table~\ref{tab:main-opensource}
reports the full $5{,}000$-item HealthBench set. For the mathematics ablations,
we randomly sample $5{,}000$ queries from DAPO-Math; the medical ablations use
the complete $4{,}865$-query LiveMedBench set. Each mathematics ablation
variant is trained for two epochs, whereas each medical ablation variant is
trained for one epoch.

Experts organize criteria as domain $\rightarrow$ sub-domain
$\rightarrow$ capability category $\rightarrow$ criterion. The mathematics
framework has $7$ domains, $24$ capability categories, and $103$ leaf
criteria; physicians independently construct the medical framework. Framework
details are in the supplementary material.

\placeablationtable

\paragraph{Benchmarks and metrics.}
Mathematics evaluation uses the competition-style AIME~24 and AIME~25 sets,
the broad-coverage MATH500~\cite{hendrycks2021math}, and the olympiad-level
OlympiadBench~\cite{he2024olympiadbench}. Medical evaluation uses HealthBench
for open-ended clinical responses~\cite{arora2025healthbench} and the
exam-style MedQA~\cite{jin2021medqa} and MedMCQA~\cite{pal2022medmcqa}.
For AIME~24 and AIME~25, we report avg@32: for each problem, we independently
sample $32$ responses and average their binary correctness.
We report each score and the macro-average within each domain.

\paragraph{Judge and reward.}
Qwen3-Max evaluates all LLM-judged benchmarks. During training,
Qwen3.7-Plus judges mathematics and a locally deployed Qwen3.6-35B-A3B judges
medicine~\cite{qwen3max,qwen3.6,qwen3.7plus}. Judges return per-rubric binary
verdicts, weight-aggregated into the scalar reward in
Equation~\ref{eq:rubric-reward}; the ORM ablation uses binary outcome
verification.

\subsection{Experimental Results}
\label{sec:experimental-results}

\subsubsection{Main Results.}
\label{sec:main-results}

Table~\ref{tab:main-opensource} compares APTER with the corresponding
unmodified model across three Qwen model generations; matched post-training
baselines and component controls are reported in Table~\ref{tab:ablation-unified}
on Qwen3-8B. APTER improves every reported
mathematics and medical benchmark, showing that its gains are not tied to a
single model generation or task format. The mathematics macro-average rises
by $14.99$, $15.86$, and $6.13$ points for Qwen2.5-7B-Instruct, Qwen3-8B,
and Qwen3.5-9B, respectively. The largest improvement is $25.82$ points on
AIME~24 for Qwen3-8B. For the already strong Qwen3.5-9B, APTER still gains
$15.50$ points on AIME~24, whereas the nearly saturated MATH500 score rises
by $2.00$ points. This pattern indicates larger benefits on difficult
competition-style reasoning tasks with greater headroom.

Medical results are likewise consistent across the three Qwen generations: the macro-average
improves by $6.63$, $5.19$, and $8.04$ points, and HealthBench gains
$17.96$, $15.42$, and $16.45$ points. The improvements on the exam-style
MedQA and MedMCQA benchmarks are generally smaller. The concentration of
gains on HealthBench is consistent with APTER's design: open-ended clinical
responses require satisfying multiple query-specific professional criteria,
whereas multiple-choice benchmarks are dominated by final-answer accuracy.
Together, the results support APTER's effectiveness across both verifiable
and expert-constrained reasoning.

\subsubsection{Ablation Experiments.}
\label{sec:ablation}

\paragraph{Ablation protocols.}
All ablations use Qwen3-8B and keep the evaluation protocol fixed within each
comparison block. The mathematics ablations use $5{,}000$ queries randomly
sampled from DAPO-Math, whereas the medical ablations use all $4{,}865$
LiveMedBench queries without subsampling. Each mathematics ablation variant is
trained for two epochs, whereas each medical ablation variant is trained for
one epoch. The medical ablations in Table~\ref{tab:ablation-unified} are
evaluated on HealthBench-500, whereas Table~\ref{tab:main-opensource} reports
the full $5{,}000$-item HealthBench set.

\noindent\textbf{Block A: Post-training components.}
SFT and RiC-SFT use identical queries, candidate responses, sample counts, and
training budgets. SFT trains on \texttt{<Query, Response>}, whereas RiC-SFT
trains on \texttt{<Query, Rubrics, Response>}. Rubric RL uses query-level
rubrics as online GRPO rewards, and APTER further adds Ada-IFT.

\noindent\textbf{Block B: Reward type.}
We hold the GRPO configuration fixed and change only the reward from binary
outcome verification to query-level rubric evaluation.

\noindent\textbf{Block C: Rubric granularity.}
We hold the data, judge, GRPO hyperparameters, and training budget fixed, and
replace query-level rubrics with five equally weighted generic criteria:
computational accuracy, logical coherence, case-splitting completeness,
variable constraint handling, and key-step coverage.

\noindent\textbf{Block D: Rubric provenance.}
For the framework-free variant, we use the question-level rubrics provided in
the original training data: SRaR for mathematics~\cite{xie2026srar} and
LiveMedBench for medicine~\cite{yan2026livemedbench}. These rubrics are
constructed independently for each question and are not routed from a shared,
expert-defined criterion framework. The expert-grounded variant instead uses
APTER rubrics routed from the corresponding domain framework, while keeping
the judge, reward formulation, and GRPO configuration fixed.

\noindent\textbf{Block E: Rubric refinement.}
We conduct two matched GRPO runs using either the raw rubrics produced by the
Consolidator or the refined rubrics produced after the Auditor--Refiner stage.
The runs use identical queries, model, judge, reward formulation, GRPO
hyperparameters, and training budget; the rubric sets average $4.82$ and
$4.09$ rubrics per question before and after refinement, respectively.

\begin{figure*}[!t]
\centering
\includegraphics[width=\textwidth]{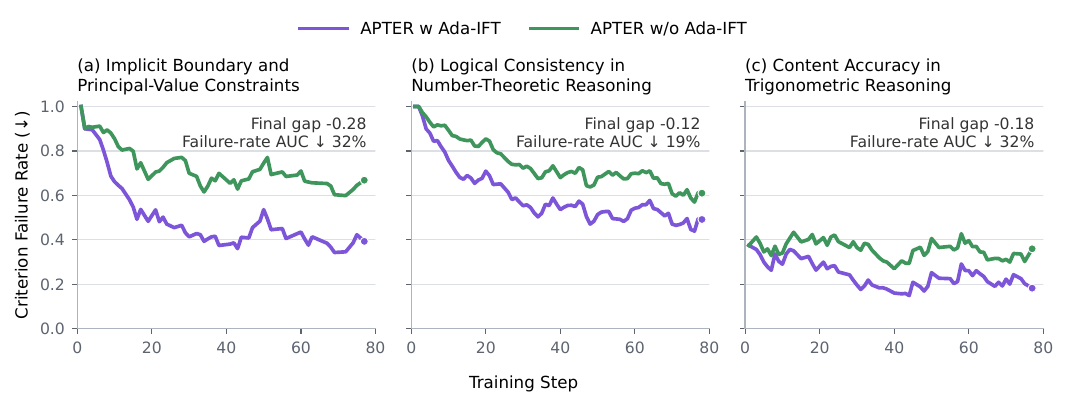}
\caption{Criterion-level repair dynamics with and without Ada-IFT.
Curves show exponential moving averages of criterion failure rates for three
randomly selected criteria; lower is better. Percentages report the
relative reduction in normalized failure-rate AUC.}
\label{fig:ada-ift-dimension-repair}
\end{figure*}

\paragraph{Post-training method ablation.}
Block~A shows that SFT alone produces mixed changes, while adding rubrics to
the supervised trajectories yields modest but consistent gains. Using rubrics
as online rewards produces the main jump: relative to RiC-SFT, Rubric
RL raises AIME~24 from $28.37$ to $41.63$ and HealthBench from $47.10$ to
$50.54$. Ada-IFT further reaches $44.60$ and $53.48$, respectively, showing
that criterion-triggered repair complements rubric-guided RL.

\paragraph{Criterion-level repair dynamics.}
Figure~\ref{fig:ada-ift-dimension-repair} examines whether Ada-IFT's aggregate
gains correspond to targeted recovery. It lowers normalized
failure-rate AUC by $32\%$, $19\%$, and $32\%$ on three randomly selected
criteria, indicating sustained rather than endpoint-only repair.

\paragraph{Reward type and rubric granularity.}
Under matched GRPO training, Block~B shows that rubric rewards outperform
binary outcome verification on all three mathematics benchmarks, with the
largest margin on AIME~24 ($37.75\!\to\!41.63$). Block~C shows that
query-level rubrics also outperform five generic criteria on every benchmark,
including HealthBench ($47.84\!\to\!50.54$), demonstrating the value of
query-specific feedback over requirements that may be too generic for the
question.

\paragraph{Rubric provenance.}
Expert grounding improves all reported results; the larger gap on AIME~24
($32.98\!\to\!41.63$) and the HealthBench gain
($47.27\!\to\!50.54$) show the value of routing rubrics from a shared expert
criterion framework rather than constructing them independently for each
question.

\paragraph{Rubric Refiner effectiveness.}
Block~E isolates the Auditor--Refiner stage. Refined rubrics yield gains of
$4.00$, $4.96$, $2.87$, and $2.36$ points on MATH500, AIME~24,
OlympiadBench, and HealthBench-500, respectively, indicating improved reward
quality rather than merely a changed rubric count. The gains are consistent
across all four benchmarks; detailed statistics are provided in the
supplementary material.

\FloatBarrier
\section{Conclusion}
\enlargethispage{2\baselineskip}

We presented APTER (Adaptive Post-Training with Expert-Grounded Rubrics), a framework that integrates structured domain knowledge into fine-grained evaluation, optimization, and capability diagnosis for specialized complex reasoning.
Its expert-grounded rubric construction component organizes domain expertise into a reusable criteria framework and instantiates relevant criteria as query-level rubrics with retained criterion provenance, without requiring per-query expert authoring or reference answers.
Its adaptive post-training component uses rubric verdicts as both optimization and criterion-level diagnostic signals, allowing Ada-IFT to aggregate recurring failures and trigger targeted supervised repair during reinforcement learning.
Experiments on mathematical reasoning and medical question answering show that the full APTER pipeline consistently improves performance across three Qwen model generations and applies to both verifiable and expert-constrained open-ended tasks.
Together, these components connect domain evaluation with targeted model improvement through expert-grounded rubrics, making post-training more domain-faithful, interpretable, and targeted.

\clearpage
\bibliographystyle{unsrtnat} 
\bibliography{main}

\clearpage
\appendix
\twocolumn[\section*{Appendix}]
\section{Method and Reproducibility Details}
\label{sec:supp-training-details}

\subsection{Computational Overhead}
\label{sec:supp-computational-overhead}

Ada-IFT shares the same judge verdicts as vanilla Rubric RL. The reward
channel aggregates rubric-level scores into scalar rewards, while the
diagnostic channel uses the same scores to compute criterion-level failure
rates and counters. The diagnostic channel therefore adds only lightweight
aggregation and indexing during ordinary RL steps. When a persistent
criterion-level deficiency triggers repair, Ada-IFT additionally invokes a
frontier model to generate candidate repair responses, verifies the
candidates, and performs a local SFT update on the accepted samples. These
extra calls occur only at triggered repair events rather than for every
rollout.

\subsection{Targeted SFT Sample Fields}
\label{sec:supp-targeted-sft-example}

When repair is triggered for criterion $c$, the frontier model receives the
query, current policy response, low-scoring query-level rubric, source
criterion and diagnostic information. Only a verified improved response
$y^\star$ is accepted. The stored supervised training tuple is
$(q,y^\star,c)$, together with provenance metadata needed for auditing. The
accepted samples are added to the criterion-indexed SFT buffer and used for
the targeted update before training returns to RL.

\subsection{Medical Training Data Composition}
\label{sec:supp-medical-data}

Table~\ref{tab:supp-medical-data} reports the exact query counts after
constructing the training files. For the II-Medical subset distributed
through RubricHub, APTER uses only the queries and constructs new
expert-grounded rubrics.

\begin{table}[htbp]
\centering
\small
\resizebox{\columnwidth}{!}{%
\begin{tabular}{lr}
\toprule
Source & Queries \\
\midrule
LiveMedBench & $4{,}865$ \\
SpeechMedDataset & $4{,}933$ \\
II-Medical (via RubricHub) & $4{,}915$ \\
\midrule
Main medical training set & $14{,}713$ \\
Medical ablation set (complete LiveMedBench) & $4{,}865$ \\
\bottomrule
\end{tabular}
}
\caption{Medical training data composition.}
\label{tab:supp-medical-data}
\end{table}

\newpage
\subsection{Expert Criteria Framework}
\label{sec:supp-criteria-framework}

\paragraph{Construction protocol.}
The mathematics framework was constructed by doctoral students with
experience in mathematical reasoning, while the medical framework was
independently constructed by a group of practicing physicians. These domain
experts first enumerate recurring capabilities that should remain meaningful
across questions. They organize these capabilities into the hierarchy domain
$\rightarrow$ sub-domain $\rightarrow$ capability
$\rightarrow$ leaf criterion. Each leaf stores a persistent criterion ID, a
scope definition, and a representative positive or boundary example. A second
review pass checks (i) coverage of common solution or clinical-response
requirements, (ii) overlap between neighboring leaves, and (iii) whether a
criterion can be instantiated as an observable, query-specific binary test.
Redundant leaves are merged and ambiguous scopes are rewritten. Approved
versions are frozen before a policy-training run; proposed changes from expert
calibration are applied only between runs, preserving the meaning of
criterion-level failure counts within a run. For compactness, the case-study
boxes later in this supplement display only the capability and leaf
criterion from a stored path; the criterion ID and provenance metadata retain
the omitted mathematics sub-domain.

\paragraph{Mathematics framework.}
Table~\ref{tab:supp-math-framework} gives the exact hierarchy stored in the
released criteria knowledge base. The seven mathematics sub-domains contain
24 capabilities and 103 leaf criteria. Some capability names recur across
sub-domains, but their definitions and examples are specialized to the
mathematical context. For example, \emph{variable constraints and boundary
handling} requires a solution to retain domain restrictions and verify
equality cases, whereas \emph{completeness of case analysis} requires an
exhaustive, non-overlapping partition for a counting problem.

\begin{table}[htbp]
\centering
\small
\resizebox{\columnwidth}{!}{%
\begin{tabular}{lrr}
\toprule
Mathematics sub-domain & Capabilities & Leaf criteria \\
\midrule
Analysis            & 4 & 16 \\
Number theory       & 4 & 16 \\
Discrete mathematics& 4 & 16 \\
Arithmetic          & 3 & 13 \\
Applied mathematics & 3 & 14 \\
Trigonometry        & 3 & 14 \\
Geometry            & 3 & 14 \\
\midrule
Total               & 24 & 103 \\
\bottomrule
\end{tabular}
}
\caption{Composition of the mathematics expert criteria framework.}
\label{tab:supp-math-framework}
\end{table}

\paragraph{Medical framework.}
The medical framework, independently designed by the participating physicians,
contains a single sub-domain, \emph{Medical Question Answering}. This sub-domain
comprises 10 capabilities and 50 criteria, listed in
Table~\ref{tab:supp-medical-framework}. These criteria are persistent
framework entries rather than query-specific rubrics. For each query, APTER
routes relevant criteria and instantiates each as an observable binary rubric.
Retaining the source criterion ID allows related failures across questions to
accumulate under the same capability.

\begin{table*}[t]
\centering
\scriptsize
\setlength{\tabcolsep}{4pt}
\renewcommand{\arraystretch}{1.08}
\begin{tabular}{p{0.22\textwidth}r p{0.66\textwidth}}
\toprule
Capability & Criteria & Criterion definitions \\
\midrule
Medical-knowledge accuracy & 7 &
Factual correctness; medical-terminology accuracy; guideline and consensus
adherence; avoidance of harmful advice; examination and test interpretation;
medication and dosage accuracy; differential-diagnosis plausibility. \\
Clinical reasoning and judgment & 5 &
Multi-factor integrated judgment; reasoning-chain clarity; severity
assessment; diagnostic and treatment prioritization; risk--benefit analysis. \\
Information completeness & 6 &
Coverage of key information; provision of alternatives; treatment-plan
completeness; follow-up and monitoring guidance; prevention and health
education; risk warnings. \\
Uncertainty handling & 5 &
Identification and communication of uncertainty; questions that reduce
uncertainty; transparent disclosure of irreducible uncertainty; balance
between a direct answer and appropriate qualification; avoidance of
overconfidence. \\
Emergency recognition and referral & 5 &
Safety-netting advice; first-aid guidance; management of panic; emergency
recognition; appropriate referral timing. \\
Context awareness and follow-up questions & 6 &
Individualized advice; proactive follow-up questions; disclosure of
assumptions and premises; multi-turn coherence; use of existing context;
avoidance of redundant questions. \\
Instruction following & 3 &
Task-requirement fulfillment; scope control; output-format compliance. \\
Communication quality and expression & 5 &
Information-density control; empathy and tone; audience adaptation;
structure and organization; emphasis of key points. \\
Health-data handling & 4 &
Accurate clinical-data extraction; identification of missing information;
preservation of data fidelity; data transformation and formatting. \\
Cultural and regional adaptation & 4 &
Awareness of regional medical-practice differences; cultural sensitivity;
language and expression adaptation; consideration of resource accessibility. \\
\midrule
Total & 50 & \\
\bottomrule
\end{tabular}
\caption{Composition of the medical expert criteria framework.}
\label{tab:supp-medical-framework}
\end{table*}

\subsection{Training Configuration}
\label{sec:supp-training-config}

For reproducibility, Table~\ref{tab:supp-hyperparams} reports the complete
Qwen3-8B recipe used by the controlled ablations and the Ada-IFT case study.
The two domain columns are separated because context length, sampling, KL
regularization, and batching differ.

\begin{table*}[htbp]
\centering
\small
\setlength{\tabcolsep}{4pt}
\renewcommand{\arraystretch}{1.12}
\begin{tabular}{p{0.29\textwidth}p{0.31\textwidth}p{0.31\textwidth}}
\toprule
Hyperparameter & Mathematics & Medicine \\
\midrule
Training data (main) & DAPO-Math-17K, 17,415 queries &
LiveMedBench + SpeechMedDataset + II-Medical, 14,713 queries \\
Ablation data (seed question) & 5,000 queries sampled from DAPO-Math-17K &
Complete LiveMedBench training set, 4,865 queries \\
Ablation budget & 2 epochs & 1 epoch \\
RL algorithm / advantage estimator & GRPO / GRPO & GRPO / GRPO \\
Reward & Weighted binary rubric reward & Weighted binary rubric reward \\
Rollouts per query $N$ & 8 & 8 \\
RL batch / PPO mini-batch & 128 / 128 & 64 / 32 \\
Optimizer & AdamW, $\beta=(0.9,0.999)$ & AdamW, $\beta=(0.9,0.999)$ \\
RL learning rate / schedule & $1{\times}10^{-6}$; constant; 10 warm-up steps &
$1{\times}10^{-6}$; constant; no warm-up \\
RL weight decay / gradient clip & 0.1 / 1.0 & 0.01 / 1.0 \\
GRPO clip (low / high / dual-$c$) & 0.2 / 0.28 / 10.0 &
0.2 / 0.2 / 3.0 \\
KL loss / coefficient & Disabled / 0 & Enabled (low-variance) / 0.001 \\
Entropy coefficient (RL) & 0 & 0 \\
Prompt / response length & 2,048 / 8,192 & 4,096 / 8,192 \\
Training sampling $(T,p,k)$ & $(1.0,1.0,-1)$ & $(0.7,0.8,20)$ \\
Validation sampling $(T,p,k)$ & $(0.7,0.8,-1)$ & $(0.7,0.8,20)$ \\
Thinking mode & Disabled & Disabled \\
Training judge & Qwen3.7-Plus & Qwen3.6-35B-A3B (local) \\
Evaluation judge & Qwen3-Max & Qwen3-Max \\
Local low-score rule & Binary verdict $s=0$ & Binary verdict $s=0$ \\
Current-step failure-rate cutoff $\rho_{\mathrm{step}}$ & 0.75 & 0.75 \\
Top-$K$ criteria / trigger count $\tau_{\mathrm{trig}}$ & 8 / 8 & 8 / 8 \\
Repair warm-up / minimum samples & 3 steps / 8 & 3 steps / 8 \\
Repair target construction & Online frontier-model generation and verification &
Online frontier-model generation and verification \\
Repair SFT batch cap / max length & 64 / 6,000 tokens &
32 / 6,000 tokens \\
Repair SFT optimizer & Independent AdamW; lr $1{\times}10^{-6}$;
$w_d=0.01$; 1 epoch; entropy $10^{-4}$ &
Independent AdamW; lr $1{\times}10^{-6}$;
$w_d=0.01$; 1 epoch; entropy $10^{-4}$ \\
Hardware & \multicolumn{2}{l}{8$\times$ NVIDIA A800-SXM4-80GB} \\
\bottomrule
\end{tabular}
\caption{Full Qwen3-8B training configuration for controlled ablations and
the Ada-IFT rollout analysis.}
\label{tab:supp-hyperparams}
\end{table*}

\noindent The ``1 epoch'' entry in the Repair SFT optimizer row refers to the
local pass over the accepted buffer samples at a triggered repair event; it
does not denote the end-to-end ablation budget.

\paragraph{Model-specific exceptions.}
For Qwen2.5-7B-Instruct, mathematics uses a 1,024-token prompt limit, a
3,072-token response limit, and PPO mini-batch size 48, whereas medicine uses
the 4,096/8,192-token limits and PPO mini-batch size 32; both use repair-batch
cap 16. Qwen3.5-9B mathematics uses the Qwen3-family context and batch sizes
but disables rollout prefix caching for its hybrid recurrent architecture.
Main mathematics runs are capped at 200 RL steps; main medical runs train for
at most two epochs and select checkpoints on the fixed HealthBench-500 subset.

\paragraph{Randomness and number of runs.}
Data-construction and SFT train--validation splitting use seed 42. The RL
dataloader uses seed 1, and the vLLM rollout engine uses seed 0 before
data-parallel worker offsets are applied. Each model--domain--method entry has
one independent random-seed run; this run count is distinct from the training
budget, which is two epochs for each mathematics ablation and one epoch for
each medical ablation. We therefore do not report across-run standard
deviations. AIME scores are \mbox{avg@32}: each problem is sampled
32 times and the reported accuracy averages those responses. Other benchmark
entries use one evaluation pass of the selected checkpoint. HealthBench-500
is a single fixed 500-sample subset rather than a newly resampled subset for
each method.

\subsection{Rubric Refiner Statistics}
\label{sec:supp-refiner-stats}

The Auditor--Refiner stage screens consolidated rubrics for redundancy,
surface bias, and boundary ambiguity. On a 5k-question subset of the
mathematics training set, only $7.0\%$ of rubric sets remain unchanged,
while the average number of criteria per question decreases from $4.82$ to
$4.09$. Table~\ref{tab:supp-refiner-stats} gives the full action statistics
underlying the main-paper analysis.

\begin{table}[H]
\centering
\small
\begin{tabular}{lr}
\toprule
Statistic & Value \\
\midrule
Rubric sets unchanged           & $7.0\%$ \\
Sets with $\geq 1$ deletion     & $61.4\%$ \\
Sets with $\geq 1$ modification & $82.4\%$ \\
Avg.\ criteria before Refiner   & $4.82$ \\
Avg.\ criteria after Refiner    & $4.09$ \\
\midrule
Overlap / redundancy issues     & $5{,}424$ \\
Surface-bias issues             & $1{,}439$ \\
Boundary-ambiguity issues       & $456$ \\
\bottomrule
\end{tabular}
\caption{Refiner action statistics on a 5k-question subset of the
mathematics training set.}
\label{tab:supp-refiner-stats}
\end{table}

\subsection{Expert Annotation Decision Flow}
\label{sec:supp-expert-annotation-flow}

This section summarizes the decision flow used by experts when annotating model responses with query-level rubrics. The goal is to avoid treating every disagreement as a single generic expert--judge mismatch. Instead, the expert first determines whether the current response can be assessed under the current criterion and rubric, then decides whether the issue comes from the model response, the judge model, or the evaluation standard itself.

\begin{figure*}[p]
\noindent\begin{minipage}{\textwidth}
\refstepcounter{algorithm}\label{alg:ada-ift}
\footnotesize
\setlength{\baselineskip}{8.8pt}
\hrule
\vspace{2pt}
\noindent\textbf{Algorithm \thealgorithm\quad Ada-IFT with criterion-indexed
targeted repair}
\vspace{2pt}
\hrule
\vspace{2pt}
\begin{tabbing}
\hspace{1.6em}\=\hspace{1.6em}\=\hspace{1.6em}\=\kill
\textbf{Input:} policy $\pi_\theta$; training set
$\mathcal{D}=\{(q,R_q)\}$; judge $J_\phi$; repair generator $F$; verifier $V$;\\
\> low-score threshold $\tau_{\mathrm{step}}$; current-step failure-rate
cutoff $\rho_{\mathrm{step}}$;\\
\> trigger count $\tau_{\mathrm{trig}}$; top-$K$ criteria; warm-up $T_0$.\\
\textbf{Initialize:} persistent counter $H_c\leftarrow 0$ for each criterion
$c$; criterion-indexed SFT buffer $\mathcal{B}_{\mathrm{sft}}\leftarrow
\varnothing$.\\[2pt]
\textbf{for} training step $t=1,2,\ldots$ \textbf{do}\\
\> Sample a query batch and draw $N$ rollouts
$\{y^{(n)}\}_{n=1}^{N}\sim\pi_\theta(\cdot\mid q)$.\\
\> \textbf{for each} $(c_i,w_i,r_i)\in R_q$ and rollout $y^{(n)}$
\textbf{do}\\
\>\> Obtain $s_i^{(n)}=J_\phi(q,y^{(n)},r_i)\in\{0,1\}$.\\
\> \textbf{end for}\\
\> Compute
$R(q,y^{(n)})=\sum_i w_i s_i^{(n)}/\sum_i w_i$ and update
$\pi_\theta$ with GRPO.\\
\> Mark local failures by $s_i^{(n)}<\tau_{\mathrm{step}}$; for every observed
criterion $c$, compute $f_c$.\\
\> Let $\mathcal{C}_t$ be the top-$K$ criteria satisfying
$f_c\geq\rho_{\mathrm{step}}$.\\
\> \textbf{for each} $c\in\mathcal{C}_t$ \textbf{do}
$H_c\leftarrow H_c+1$.\\
\> \textbf{if} $t>T_0$ and some $H_c\geq\tau_{\mathrm{trig}}$
\textbf{then}\\
\>\> Select the highest-count eligible criterion $c^\star$.\\
\>\> For representative failures of $c^\star$, generate candidates\\
\>\> \quad $y^\star\sim F(q,y,r,c^\star,\text{diagnostic information})$.\\
\>\> Verify candidates with $V$ and add accepted
$(q,y^\star,c^\star)$ to $\mathcal{B}_{\mathrm{sft}}$.\\
\>\> \textbf{if} at least the minimum number of samples is available
\textbf{then}\\
\>\>\> Apply one targeted SFT update on the buffered samples and reset
$H_{c^\star}\leftarrow 0$.\\
\>\> \textbf{end if}\\
\> \textbf{end if}\\
\textbf{end for}
\end{tabbing}
\vspace{-5pt}
\hrule
\end{minipage}

\par\nopagebreak[4]
\vspace{3pt}
\noindent\begin{minipage}{\textwidth}
\centering
\captionsetup{type=figure,skip=3pt}
\includegraphics[width=0.95\textwidth]{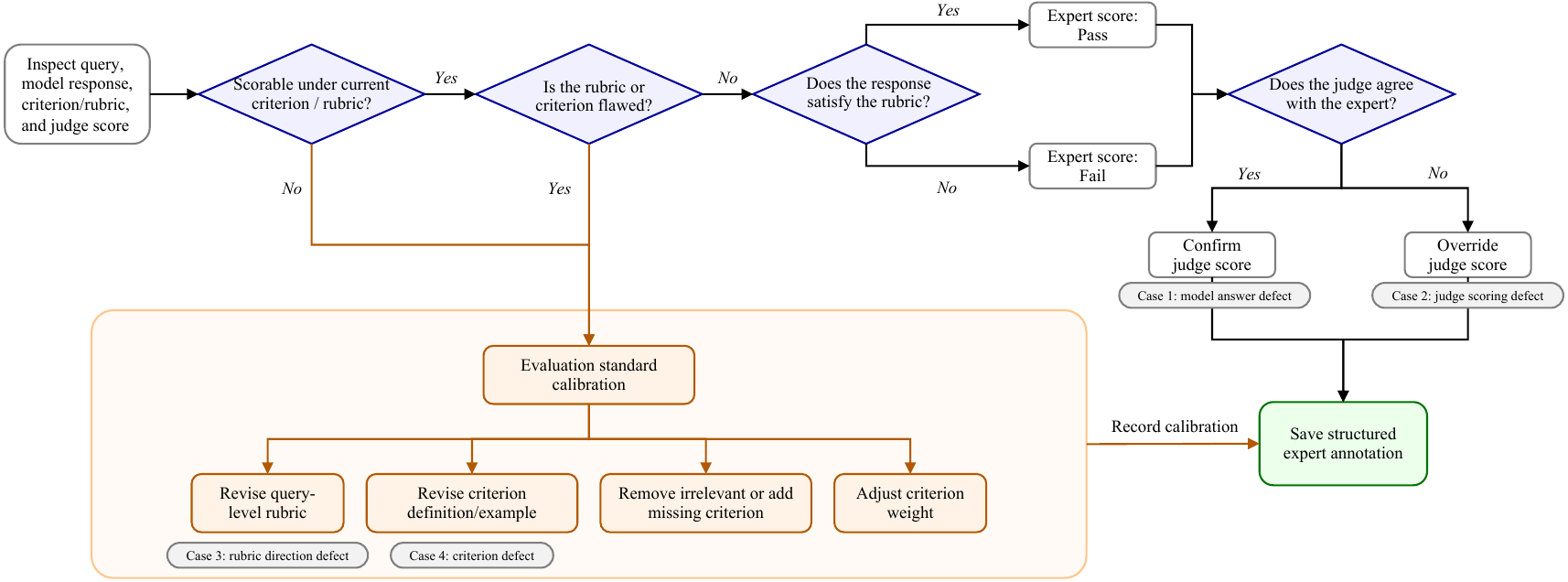}
\captionof{figure}{Expert annotation decision flow. Invalid evaluation
standards enter the shared calibration module; valid standards proceed to
expert--judge score comparison.}
\label{fig:expert-annotation-flow}
\end{minipage}
\end{figure*}

Figure~\ref{fig:expert-annotation-flow} shows the decision process. The key design choice is that both unscorable cases and flawed-standard cases enter the same \emph{evaluation standard calibration} module. This is because an unscorable response usually indicates that the current criterion or rubric is irrelevant, underspecified, or otherwise unsuitable for the item. APTER therefore exposes a shared set of standard-calibration actions: revise the query-level rubric, revise the criterion definition or example, remove an irrelevant criterion or add a missing one, and adjust the criterion weight.

In this flow, score confirmation and score overriding are reserved for cases
where the evaluation standard is valid. When the standard itself is flawed,
the expert records a structured calibration action rather than merely changing
the score. The resulting annotation can later support judge calibration,
rubric-generation repair, criterion-framework refinement, or weight adjustment.

\clearpage
\raggedbottom

\clearpage
\onecolumn
\section{Prompt Templates}
\label{sec:supp-prompts}

This section lists the prompt templates used at each stage of the APTER
pipeline. We report them here to make the data-construction and
post-training process reproducible. For conciseness, only the mathematics
versions are provided as representative examples. Placeholders such as \texttt{\{query\}},
\texttt{\{criterion\}}, \texttt{\{rubric\}}, and \texttt{\{response\}}
denote the fields that are filled in at runtime. The exact prompt bodies
are provided below. In these implementation prompts, \emph{dimension} is the
serialized name of a selected framework criterion, not an additional level
in the expert-criteria hierarchy.

\subsection{Query-level Routing Prompt}
\label{sec:supp-prompt-routing}

Given a query and the expert criteria framework, the routing prompt
selects the relevant candidate criteria and assigns each a weight
(see the rubric-generation section of the main paper).
\begin{AIbox}[breakable]{Query-level Routing Prompt}
\begin{lstlisting}
# 1. Role
You are a top-tier AI evaluation architect with exceptional systems thinking and analytical ability. Your task is the initial design of an evaluation framework for solutions to complex mathematics problems.

# 2. Core Mission
Act as the "chief planner" of the evaluation framework and carry out a highly focused FIRST-STAGE task. Analyze the user-provided [query] and, based on it, accomplish two things:
1. Dimension selection: You MUST select ONLY from the [candidate dimension list] of the specified mathematics domain the subset of evaluation dimensions most relevant to this [query]. Do NOT create, modify, combine, or rename any dimension.
2. Weight assignment and ranking: For each selected dimension, assign an importance weight in [0, 10] according to how important it is for judging the quality of a response to this [query].

# 3. Instructions and Procedure
Follow this reasoning chain strictly:
1. Analyze the query in depth: identify the mathematical concepts, solution methods, and evaluation needs it involves.
2. Select and weight:
   a. Quickly scan the user-provided [candidate dimension list].
   b. Based on your analysis, SELECT 5-10 of the most relevant dimensions from the list.
   c. Assign each selected dimension a weight in [0, 10]:
      - High (9-10): core, decisive criteria for this response, possibly "veto" items.
      - Medium (6-8): important but non-core checkpoints.
      - Low (1-5): relevant but secondary checkpoints, or "bonus" items.
3. [Reinforced] Final check: before producing output, verify that every "dimension" value in the final JSON list comes VERBATIM from the user-provided list.
4. Rank and format: sort all selected, weighted dimensions in descending order of weight and output strictly in the required format.

# 4. Required Output Format
Output strictly as a Python list of JSON objects, with no other explanatory text or code-block markers:
{"dimensions": [
  {"dimension": "name of the highest-weight dimension", "weight": score},
  {"dimension": "name of the second most important dimension", "weight": score},
  ...
]}

# 5. Example
Suppose the problem belongs to the "Analysis" domain; a possible result:
{"dimensions": [
  {"dimension": "Analytical rigor", "weight": 10},
  {"dimension": "Quantitative computation accuracy", "weight": 9}
]}
\end{lstlisting}
\end{AIbox}

\subsection{Multi-role Rubric Generation Prompts}
\label{sec:supp-prompt-generation}

APTER instantiates the selected criteria into query-level rubrics
through a multi-role process (Analyst, Consolidator, Auditor, Refiner).
The prompt for each role is given below.

\begin{AIbox}[breakable]{Analyst Prompt}
\begin{lstlisting}
# Role
You are a senior mathematics curriculum researcher and assessment-design expert with decades of experience setting and grading exams. Your strength: given a problem, you quickly see through its core testing points, key steps, and where most people go wrong.

# Core Mission
You will receive a mathematics problem and an expert evaluation framework (a set of candidate dimensions). Produce a DEEP, FREE-FORM analysis of the problem. Your analysis will be used by another expert to design scoring dimensions, so quality is critical.
Write your analysis freely in natural language; do NOT output any JSON or structured format. Develop your thinking fully.

# Guiding Questions
The following are directions worth considering; you need not answer each one, but let them inspire a deep analysis:
- How would you approach solving this problem? Try to work it through completely and give your final answer.
- What are the key intermediate conclusions, i.e. steps such that if one is wrong, everything after it is necessarily wrong?
- Are there "looks-valid-but-is-a-trap" solution paths? What trap might the problem setter be hinting at?
- Which mathematical concepts or theorems are the real pillars for solving this problem? Someone who does not understand them cannot get it right.
- Conversely: what distinguishes someone who can truly solve this problem from someone who "writes a lot but misses the point"? Which dimension best separates the two?
- Among the candidate dimensions in the framework, which are truly discriminative FOR THIS SPECIFIC PROBLEM, and which are irrelevant here?
- If you could pick only 3 dimensions to judge an answer, which 3 would you pick, and why?

# Key Principles
- The dimensions you recommend MUST be crucial to correctly solving this problem. If a dimension does not affect answer correctness, it is not worth being a scoring criterion. You may pick 5-7 dimensions.
- Every number, formula, or theorem in your analysis must be correct. If unsure about a conclusion, explicitly flag your uncertainty rather than fabricating a plausible-sounding claim.
- If the problem admits multiple valid solution paths, identify them and note this in your analysis.
\end{lstlisting}
\end{AIbox}

\begin{AIbox}[breakable]{Consolidator Prompt}
\begin{lstlisting}
# Role
You are a rubric-synthesis expert. Your task is to distill high-quality scoring dimensions from multiple independent analysis reports.

# Core Mission
You will receive a mathematics problem, multiple independent in-depth analysis reports from different analysts, and the expert evaluation framework. Synthesize all information into a final list of scoring dimensions.

# Workflow
## Step 1: Cross-validate facts
Compare the concrete numbers, intermediate conclusions, and theorem citations across the reports:
- If most reports agree, adopt the majority view.
- If they disagree, prefer the report with the most complete and rigorous derivation, and verify it repeatedly yourself.
- If one report makes a mathematical claim no other report mentions, treat it with caution: do not adopt it unless its derivation is convincing.

## Step 2: Identify the truly critical dimensions
From the analyses, distill the dimensions most critical to CORRECTLY solving this problem:
- Prefer dimensions emphasized by multiple reports; consensus among independent analysts is a strong signal.
- Focus on the "get-it-wrong-and-all-is-lost" key steps and concepts.
- Discard generic dimensions that are "correct" but non-discriminative for this problem.

## Step 3: Check logical consistency
Before final output, check for contradictions among the selected dimensions:
- Are there two dimensions whose requirements cannot logically hold simultaneously?
- If the problem admits multiple valid solution paths, the criterion must not hard-code one path as the only correct way.
- The dimensions must be mutually independent: do not test the same computation step or logical point in two dimensions.

# Output Guidelines
1. Precise selection: usually around 5 dimensions, but analyze the specific problem, pick more if needed, fewer if fewer suffice; never pad the count.
2. Weighting: 1-10; larger means more central to correct solving.
3. Criteria must be concrete: include the specific numbers, formulas, and step anchors of this problem; no vague generalities.
4. Binary objective judgment: each criterion must support a clear "satisfied / not satisfied" verdict; no fuzzy wording.
5. Dimension names: must be chosen from the candidate dimensions of the expert framework, using their full names.
\end{lstlisting}
\end{AIbox}

\begin{AIbox}[breakable]{Auditor Prompt}
\begin{lstlisting}
# Role
You are a senior rubric quality-control expert, responsible for auditing and diagnosing the reliability, atomicity, and discriminativeness of automatically generated scoring criteria.

# Core Mission
Your task is highly focused: given a [Query] and a set of [candidate Rubrics] generated for it (a list[dict], each element a dimension criterion), act as a strict examiner. Check for logical overlap among the rubrics, diagnose the internal defects of each rubric, and finally route their "dimension names" into three groups: Keep, Modify, or Delete.

# Required Input
1. [Query]: the original query text or task instruction to be evaluated.
2. [Candidate Rubrics]: the list of scoring criteria generated for the query (JSON list[dict]; each element contains at least dimension / weight / criterion).

# Instructions
1. Atomicity and overlap detection: high-quality criteria must be atomic. Carefully compare all candidate criteria and find those that test the same step, repeat a checkpoint, or logically cross over heavily. Group the overlapping dimension names into the same list.
2. Per-criterion defect diagnosis: for each dimension's criterion, check for the following fatal defects and state the issue concisely (write "no issue" if flawless):
   - Non-atomic: one criterion mixes multiple points that need independent scoring.
   - Off-instruction: generic, not strictly derived from the specific information in this Query.
   - Surface-bias tendency: over-focused on surface features such as length or formatting, easily misled by a model's verbosity, lacking substantive logical checking.
   - Boundary ambiguity: contains subjective or ambiguous wording, lacking an objective decision anchor.
3. Action categorization: based on the diagnosis, route every input dimension name into exactly one of three groups:
   - Keep: perfectly fits the Query, unambiguous, non-overlapping, and atomic; ready for use.
   - Modify: direction is correct but has boundary ambiguity, compound dimensions, or slight overlap (e.g. several dimensions checking the same addition); needs later polishing or merging/splitting.
   - Delete: generic filler with no discriminativeness, severely off-Query, or fully redundant with a Kept item.
\end{lstlisting}
\end{AIbox}

\begin{AIbox}[breakable]{Refiner Prompt}
\begin{lstlisting}
# Role
You are a top-tier rubric optimization and reconstruction expert. Your specialty is targeted repair of flawed draft criteria, turning them into high-quality, atomic rubrics strictly aligned with human-expert standards, thereby defending against surface-level biases in LLM evaluation.

# Core Mission
You receive the original [Query], the [original Rubrics], and the full [diagnostic feedback] (the exact Keep / Modify / Delete lists and defect notes). Act as an assembly-line inspector: firmly discard items to delete, directly inherit items to keep, and deeply reconstruct items to modify and those with overlaps, finally outputting a flawless set of rubrics.

# Required Input
1. [Query]: the original query text or task instruction to be evaluated.
2. [Original Rubrics]: the initially generated rubric set (JSON, with each dimension's weight and criterion).
3. [Diagnostic feedback]: the detailed diagnosis from the quality-check module, including:
   - keep_rubrics: dimension names recommended to keep as-is.
   - delete_rubrics: dimension names recommended to delete.
   - modify_rubrics: dimension names recommended to modify.
   - rubric_issues: specific defect notes (for the modify items).
   - overlaps: groups of logically overlapping dimensions.

# Refinement & Synthesis Guidelines (must follow strictly)
1. Naming invariants:
   - The dimension names in keep_rubrics / modify_rubrics / delete_rubrics must be echoed and referenced verbatim; no rewriting, abbreviation, keeping only the second-level name, or synonym translation.
   - These names are already in "level-1 dimension - level-2 dimension" full-name format; keep that full-name format unchanged throughout.
2. Execute categorization precisely:
   - Absolute delete: any dimension in delete_rubrics is erased from your reasoning and final output; never keep it.
   - Keep as-is: any dimension in keep_rubrics has its name, weight, and criterion copied verbatim into the final output.
   - Targeted modify: only dimensions in modify_rubrics are re-polished and decomposed according to the defects in rubric_issues (e.g. compound dimension, boundary ambiguity, off-instruction).
3. Resolve overlaps: inspect the overlaps list. If dimensions A and B overlap, you must decide: either MERGE them into a single new dimension covering that checkpoint (delete the original A and B, and the new dimension's name must be chosen from A or B, do not invent a new name), or strictly revise their criteria to draw a clear boundary so they test entirely independent steps.
4. Enforce atomicity & mitigate bias: each reconstructed modify item must test one indivisible logical node. The criterion must be derived strictly from the specific information in the Query, must include concrete numbers, formulas, or step anchors as objective decision bases, and must eliminate subjective vague wording such as "basically correct" or "logically clear" that is easily misled by verbosity.
\end{lstlisting}
\end{AIbox}

\subsection{Rubric Judging Prompt}
\label{sec:supp-prompt-judge}

The implementation batches all rubrics associated with one response. The
following is the medical judging template; the mathematics variant retains the
same input fields and Boolean output contract but removes explanations and
asks the judge to ignore superficial formatting differences.
\begin{AIbox}[breakable]{Rubric Judging Prompt}
\begin{lstlisting}
Your job is to look at a conversation and a list of rubric items, and
score the last assistant response on how well it follows each item.

# Conversation
{conversation}

# Rubric items
{rubric_items}

# Instructions
Return a JSON object with one field, "results", whose value is a list
of objects in the same order as the rubric items. Each object contains
"explanation" and "criteria_met".
- "criteria_met" is Boolean. Return true only if the response satisfies
  the entire rubric item; if any required part is missing or incorrect,
  return false.
- Text following "such as", "for example", or "including" is
  illustrative rather than an exhaustive checklist.
- If a rubric describes an undesirable behavior with negative weight,
  judge whether that behavior is present; do not invert the verdict.
- The result list must contain exactly one object per rubric item.

Return only the JSON object. Do not include other text.
\end{lstlisting}
\end{AIbox}

\begin{AIbox}[breakable]{Mathematics Judge Output Addendum}
\begin{lstlisting}
You are a strict grader for competition math. Judge each rubric item
independently against the student's solution. Ignore superficial
formatting differences such as boxed versus plain answers, LaTeX versus
plain text, units, trailing zeros, and mathematically equivalent forms.

Return ONLY:
{"results":[true,false,...]}
The array must contain exactly one Boolean per rubric item, in order.
Do not explain and do not output Markdown.
\end{lstlisting}
\end{AIbox}

\subsection{Candidate Response Generation Prompt}
\label{sec:supp-prompt-candidate}

Candidate generation receives the original query and its complete weighted
rubric set. Multiple sampled responses are judged with the preceding prompt;
the highest-scoring verified response becomes the supervised target. The
runtime user-message template is:
\begin{AIbox}[breakable]{Candidate Response Generation Prompt}
\begin{lstlisting}
# Question
{query}

# Evaluation Criteria
{weighted_query_level_rubrics}

Provide a complete answer to the question. Address the evaluation
criteria without mentioning, quoting, or revealing that a rubric was
provided. Use the same language as the question. For mathematics,
show a rigorous derivation and clearly mark the final answer.
\end{lstlisting}
\end{AIbox}

\subsection{Targeted Repair Prompt (Ada-IFT)}
\label{sec:supp-prompt-repair}

The following prompt is used after Ada-IFT triggers a criterion-indexed repair
event during policy training. When a diagnostic report is available, the
failed variant is used; a shorter variant omits the report and negative
example when those fields are unavailable.
\begin{AIbox}[breakable]{Targeted Repair Prompt}
\begin{lstlisting}
## Role
You are a senior domain expert with exceptionally rigorous reasoning.
Produce a high-quality solution that excels on one scoring dimension
where the policy currently struggles.

## Task
You receive a query, a scoring dimension with its binary criterion, a
diagnostic report of common errors, and a typical failing response.
Solve the query from scratch and produce a complete exemplary response.

## Critical Rules
1. The response must particularly excel on the specified dimension and
   strictly satisfy its binary criterion.
2. Deliberately avoid the error patterns in the diagnostic report.
3. Do not mention or critique the failing response.
4. Maintain domain correctness and use the same language as the query.
5. Give a clear final answer or recommendation.

## Input
### Scoring Dimension
- Dimension Name: {criterion_id}
- Category: {capability_category}
- Binary Criterion: {query_level_rubric}

### Query
{query}

### Diagnostic Report
- Policy Pass Rate: {pass_rate}
- Common Error Patterns: {common_errors}
- Failure Summary: {failure_summary}

### Typical Failing Response
{policy_response}
\end{lstlisting}
\end{AIbox}

\subsection{RiC-SFT Sample Format}
\label{sec:supp-ric-sft-format}

Rubric-in-CoT SFT (RiC-SFT) uses query-level rubrics as lightweight reasoning
guidance before answer generation. A training sample can be formatted as
follows:
\begin{lstlisting}
query: ...
response:
<think>
I should first consider the rubrics for this query before answering.
The response should satisfy the following criteria:
1. ...
2. ...
Based on these rubrics, I need to ...
</think>
...
\end{lstlisting}

\section{Case Studies}
\label{sec:supp-expert-annotation-case-study}

\subsection{Expert Annotation and Calibration}

This supplementary section presents a small expert-annotation case study. The purpose of this case study is not to evaluate mathematical difficulty itself, but to illustrate how APTER supports expert calibration at multiple levels of the evaluation pipeline. In particular, the cases demonstrate how experts handle model-answer defects, judge-model scoring defects, rubric-direction defects, and criterion-level defects.

\subsubsection{Case Study Design}
\label{sec:supp-case-study-design}

The case study contains four mathematics items. Each item is designed to represent a distinct failure mode in rubric-based evaluation. For each item, APTER presents the query, model response, criteria and query-level rubrics to be annotated, per-rubric binary verdicts, their weighted aggregate, expert scores, and expert resolution actions. This makes the annotation process auditable: readers can inspect not only the final score, but also which part of the evaluation pipeline required expert intervention.

Table~\ref{tab:supp-expert-cases-overview} summarizes the four cases.

\begin{table}[H]
\centering
\small
\begin{tabular}{p{0.06\textwidth} p{0.20\textwidth} p{0.10\textwidth} p{0.10\textwidth} p{0.38\textwidth}}
\hline
\textbf{Case} & \textbf{Issue Type} & \textbf{Judge Aggregate} & \textbf{Expert Score} & \textbf{Expert Calibration Action} \\
\hline
1 & Model answer defect & 0 & 0 & Confirm the judge-model low score because the model answer is genuinely incorrect. \\
2 & Judge scoring defect & 0.5 & 1 & Override the judge-model score while keeping the rubric unchanged. \\
3 & Rubric direction defect & 1 & 0 & Mark the rubric as flawed and rewrite it toward the correct proof-oriented evaluation direction. \\
4 & Criterion defect & 0.7 & 1 & Mark a single criterion as flawed and revise its definition and example, while keeping the overall rubric direction. \\
\hline
\end{tabular}
\caption{Overview of the expert-annotation case study. Each case corresponds to a different calibration level in APTER.}
\label{tab:supp-expert-cases-overview}
\end{table}

\subsubsection{Case 1: Model Answer Defect}
\label{sec:supp-case-model-answer-defect}
\mbox{}\par\vspace*{10pt}

\begin{AIbox}[breakable,before skip=12pt,top=12pt,
  attach boxed title to top left={yshift=-4mm,xshift=2mm}]
  {Case 1 --- Judge 0 / Expert 0 --- Model Answer Defect}
\begin{lstlisting}
[Query]
A jacket originally costs USD 80. It is discounted by 25%, and then an 8%
sales tax is applied to the discounted price. What is the final price?

[Model Response]
The model computes the 25% discount correctly and obtains USD 60 as the
discounted price. However, it then states that an 8% tax means adding
USD 8, and reports the final answer as USD 68.

[Criteria & Rubrics to Annotate]
R1  (Foundational Mathematical Modeling -- Constraint Extraction &
     Transformation)   weight 5
    Judge: 0    Expert: 0
    The response must identify that the 8% sales tax is applied to the
    discounted price, not to the original price.

R2  (Foundational Mathematical Modeling -- Calculation Accuracy)  weight 5
    Judge: 0    Expert: 0
    The response must compute the tax as 0.08 * 60 = 4.80 and report the
    final price as USD 64.80.

[Expert Calibration Action]
Confirm the judge-model low score. The response applies the tax
incorrectly and reports the wrong final price, so the defect lies in the
model answer, not in the judge model or the rubric. This is the only pure
agreement case in the set.
\end{lstlisting}
\end{AIbox}

\paragraph{Analysis.}
This case demonstrates model-answer validation: the judge and the expert agree, and the low score reflects a genuine error in the response rather than any flaw in the evaluation standard.

\subsubsection{Case 2: Judge Scoring Defect}
\label{sec:supp-case-judge-scoring-defect}
\mbox{}\par\vspace*{10pt}

\begin{AIbox}[breakable,before skip=12pt,top=12pt,
  attach boxed title to top left={yshift=-4mm,xshift=2mm}]
  {Case 2 --- Judge 0.5 / Expert 1 --- Judge Scoring Defect}
\begin{lstlisting}
[Query]
Solve for x:  2(x - 3) = 10.

[Model Response]
The model divides both sides by 2, obtains x - 3 = 5, adds 3 to both
sides, and reports the final answer x = 8.

[Criteria & Rubrics to Annotate]
R1  (Foundational Cognition & Knowledge -- Content & Method Accuracy)
    weight 5
    Judge: 1    Expert: 1
    The response must use valid algebraic operations to solve the equation
    and obtain x = 8.

R2  (Mathematical Communication -- Step Clarity & Granularity)  weight 5
    Judge: 0    Expert: 1
    The response must include enough intermediate reasoning to verify the
    solution. Expanding 2(x - 3) into 2x - 6 is optional.

[Expert Calibration Action]
Override the judge-model score while keeping the rubric unchanged. The
response is mathematically correct, and dividing by 2 is a valid and
sufficiently clear transformation; the judge applied the step-clarity
rubric too rigidly by treating the missing expansion as a failure.
\end{lstlisting}
\end{AIbox}

\paragraph{Analysis.}
This case demonstrates judge-score correction: the rubric itself is reasonable, so the expert corrects the automated score without touching the evaluation standard.

\subsubsection{Case 3: Rubric Direction Defect}
\label{sec:supp-case-rubric-direction-defect}
\mbox{}\par\vspace*{10pt}

\begin{AIbox}[breakable,before skip=12pt,top=12pt,
  attach boxed title to top left={yshift=-4mm,xshift=2mm}]
  {Case 3 --- Judge 1 / Expert 0 --- Rubric Direction Defect}
\begin{lstlisting}
[Query]
Prove that for every integer n, if n^2 is even, then n is even.

[Model Response]
The model checks examples: 2^2 = 4, 4^2 = 16, and 6^2 = 36. It then states
that these examples show that when n^2 is even, n is even.

[Criteria & Rubrics to Annotate]
R1  (Reasoning & Synthesis -- Method & Induction Rigor)  weight 5
    Judge: 1    Expert: 0
    The generated rubric is flawed: it only asks for several correct
    examples rather than a proof for every integer.

R2  (Mathematical Communication -- Method Description Rigor)  weight 5
    Judge: 1    Expert: 0
    The generated rubric is flawed: it accepts stating the correct
    conclusion without requiring a logically valid proof method, such as
    contrapositive reasoning or parity cases.

[Judge vs. Expert]
Under the flawed rubric the judge assigns 1 to both rubrics (examples are
given and the conclusion stated). The expert assigns 0: examples cannot
establish a universal statement.

[Expert Calibration Action]
Mark the rubric as flawed and select the rubric-revision path. The revised
rubric should identify the universal proof obligation, require a valid
contrapositive or parity argument, show that every odd integer has an odd
square, and make the conclusion follow logically for all integers.
Examples may support intuition but cannot serve as sufficient evidence.
\end{lstlisting}
\end{AIbox}

\paragraph{Analysis.}
This case demonstrates rubric-level redirection: the failure is not a wrong score on a single criterion but a wrong evaluation \emph{direction} for the task type, so the whole query-level rubric must be rewritten.

\subsubsection{Case 4: Criterion Defect}
\label{sec:supp-case-criterion-defect}
\mbox{}\par\vspace*{10pt}

\begin{AIbox}[breakable,before skip=12pt,top=12pt,
  attach boxed title to top left={yshift=-4mm,xshift=2mm}]
  {Case 4 --- Judge 0.7 / Expert 1 --- Criterion Defect}
\begin{lstlisting}
[Query]
A recipe uses 3 cups of flour for every 2 cups of sugar. If you use 9 cups
of flour, how many cups of sugar should you use?

[Model Response]
The model states that the flour amount is multiplied by 3, from 3 cups to
9 cups. It then multiplies the sugar amount by the same factor:
2 * 3 = 6. The final answer is 6 cups of sugar.

[Criteria & Rubrics to Annotate]
R1  (Reasoning & Synthesis -- Rule & Relationship Relevance)  weight 4
    Judge: 1    Expert: 1
    The response must identify that the flour and sugar amounts should be
    scaled by the same factor.

R2  (Foundational Arithmetic Cognition -- Calculation Accuracy)  weight 3
    Judge: 1    Expert: 1
    The response must correctly compute 2 * 3 = 6 and give 6 cups of sugar.

R3  (Mathematical Communication -- Step Clarity & Granularity)  weight 3
    Judge: 0    Expert: 1
    The generated criterion is defective for this case study: it requires
    at least three calculation steps to receive full credit.

[Expert Calibration Action]
Mark the single criterion R3 as flawed and select the criterion-revision
path. Replace the fixed step-count requirement with a sufficiency-based
criterion: the response should provide enough intermediate reasoning to
make the proportional scaling verifiable, but the number of calculation
steps is not fixed. A valid example: the flour amount is multiplied by 3,
so the sugar amount is also multiplied by 3, giving 2 * 3 = 6 cups.
\end{lstlisting}
\end{AIbox}

\paragraph{Analysis.}
This case demonstrates criterion-level repair. Unlike Case~3, the overall rubric direction is sound; only one criterion's operational standard (a fixed step count) is flawed and needs revision, while the rest of the rubric is kept intact.

\subsubsection{Discussion}
\label{sec:supp-case-discussion}

These four cases show that expert annotation in APTER is not limited to assigning a final correctness label. Instead, experts can intervene at multiple levels:

\begin{itemize}
    \item \textbf{Model-answer level}: confirm that a low score is caused by a genuine response error.
    \item \textbf{Judge-model level}: correct an automated scoring error without changing the rubric.
    \item \textbf{Rubric level}: rewrite a query-level rubric when its evaluation direction is wrong.
    \item \textbf{Criterion level}: revise an individual criterion when its wording or operational standard is flawed.
\end{itemize}

This multi-level calibration mechanism is important for making rubric-based supervision reliable. It prevents all disagreements from being collapsed into a single ``expert versus judge'' label, and instead records the root cause of each disagreement as structured supervision for future rubric generation, judge calibration, and expert criteria framework refinement.

\subsection{Effectiveness of Process-based Rubrics in RL}
\label{sec:supp-rl-rubric-case-study}

While Section~\ref{sec:supp-expert-annotation-case-study} shows how experts
calibrate rubrics during data construction, this section presents three
\emph{real} rollouts from the Qwen3-8B RL training logs
(step~$20$ of~$200$). The cases were selected to expose three qualitatively
different reward outcomes: a correct final answer supported by unsound
reasoning, a failed trajectory that produces a criterion-level repair signal,
and a fully correct solution. Each box contains the problem, query-level
rubrics, per-criterion judge verdicts, weighted reward, and complete original
response, translated into English and lightly reformatted. All three rollouts
use the same generation prompt.

\begin{AIbox}[breakable]{Common Generation Prompt (shared by all queries)}
\begin{lstlisting}
Solve the following math problem step by step. Follow these formatting rules:
1. Steps: Break your solution into multiple clear steps. Begin each step with "### Step N:" (e.g., ### Step 1:, ### Step 2:, ### Step 3:).
2. Final Answer: End your response with the answer inside \boxed{}.

Problem: {problem}
\end{lstlisting}
\end{AIbox}

Table~\ref{tab:supp-rl-cases-overview} summarizes the three cases. The key
contrast is between outcome correctness and process quality: Case~1 receives
zero despite matching the reference answer, Case~2 exposes a persistent
capability failure, and Case~3 receives full credit only after satisfying all
required reasoning checks.

\begin{table}[H]
\centering
\small
\setlength{\tabcolsep}{5pt}

\begin{tabular}{c c c p{0.52\textwidth}}
\toprule
\textbf{Case} & \textbf{Rubric Score} & \textbf{Answer} & \textbf{What the rubric reveals} \\
\midrule
1 & $0.000$ & Correct & Reward hacking: a correct answer reached through an unsound approximation with no error control. \\
2 & $0.000$ & Wrong   & Reasoning collapse: a hallucinated ``known result''; the persistent zero trips the Ada-IFT trigger. \\
3 & $1.000$ & Correct & Clean, fully rigorous solution earning full reward. \\
\bottomrule
\end{tabular}
\caption{Overview of three representative RL rollout cases (Qwen3-8B,
step~20). ``Answer'' denotes whether the final boxed answer matches the
reference.}
\label{tab:supp-rl-cases-overview}
\end{table}

\subsubsection{Case 1: Reward Hacking --- Correct Answer, Unsound Method}
\label{sec:supp-case-reward-hacking}
\mbox{}\par\vspace*{10pt}

\begin{AIbox}[breakable,before skip=12pt,top=12pt,
  attach boxed title to top left={yshift=-4mm,xshift=2mm}]
  {Case 1 --- Rubric Score: 0.000 --- Final Answer: Correct}
\begin{lstlisting}
====================================
Reference Answer: 8   |   Rubric Max: 34 pts   |   Response Length: 1835 chars
====================================

[Problem]
Given the sequence {a_n} with a_n = sqrt( (2n - 1) / (4n^2 + 1) ), let
S_n be the sum of its first n terms. Find the integer closest to
S_128 - S_32.

[Rubric Criteria & Judge Scores]
D1  (Reasoning & Synthesis / Strategy Planning & Path Optimization)
    weight 9   ->   Score 0.0  [FAIL]
    Chooses an effective summation-approximation strategy (e.g., a
    telescoping bound from the asymptotic a_n ~ 1/sqrt(2n), or an
    integral approximation with explicit remainder control) and sets an
    error-control target confining S_128 - S_32 to an interval of length
    <= 1, such as (7.5, 8.5).

D2  (Analysis Tools / Inequality Construction & Application)
    weight 10  ->   Score 0.0  [FAIL]
    Constructs, for a_n = sqrt((2n-1)/(4n^2+1)), a two-sided inequality
    f(n) < a_n < g(n) holding strictly for n in [33, 128], where f and g
    are summable functions (e.g., radical differences of the form
    sqrt(2n + c1) - sqrt(2n + c2)).

D3  (Foundational Cognition & Knowledge / Quantitative Computation Accuracy)
    weight 8   ->   Score 0.0  [FAIL]
    When summing f and g, the telescoped boundary terms (n=33 and n=128)
    are substituted correctly, and the resulting bounds on S_128 - S_32
    both fall strictly within (7.5, 8.5), yielding 8 as the closest
    integer.

D4  (Foundational Cognition & Knowledge / Analytical Logical Consistency)
    weight 7   ->   Score 0.0  [FAIL]
    In the algebraic manipulation of the radical inequalities (cross-
    multiplication, squaring), non-negativity preconditions are verified
    and the bounding direction stays consistent, with no logical jumps or
    reversed inequality signs.

--> Weighted rubric total: 0.000

[Model Response (full)]
We know the general term of the sequence {a_n} is:

$$ a_n = \sqrt{\frac{2n - 1}{4n^2 + 1}} $$

We need S_128 - S_32, where S_n is the sum of the first n terms:

$$ S_n = a_1 + a_2 + \cdots + a_n $$

So we want:

$$ S_{128} - S_{32} = a_{33} + a_{34} + \cdots + a_{128} $$

### Step 1: Rewrite as a summation
$$ S_{128} - S_{32} = \sum_{n=33}^{128} a_n = \sum_{n=33}^{128} \sqrt{\frac{2n - 1}{4n^2 + 1}} $$

### Step 2: Simplify the general term
Look at the relation between the denominator 4n^2 + 1 and the numerator
2n - 1:
$$ 4n^2 + 1 = (2n + 1)(2n - 1) + 2 $$
This is not obviously helpful. Note also
$$ \frac{2n - 1}{4n^2 + 1} = \frac{2n - 1}{(2n - 1)(2n + 1) + 2} $$
and 4n^2 + 1 = (2n)^2 + 1, which still does not simplify directly, so we
try a simpler method.

### Step 3: Approximate the general term
Since
$$ a_n = \sqrt{\frac{2n-1}{(2n)^2+1}} \approx \sqrt{\frac{2n}{4n^2}} = \sqrt{\frac{1}{2n}} = \frac{1}{\sqrt{2n}} $$
this approximates a_n well for large n:
$$ a_n \approx \frac{1}{\sqrt{2n}} $$

### Step 4: Estimate the sum
$$ S_{128} - S_{32} \approx \sum_{n=33}^{128} \frac{1}{\sqrt{2n}} = \frac{1}{\sqrt{2}} \sum_{n=33}^{128} \frac{1}{\sqrt{n}} $$
Approximate the inverse-square-root sum by an integral:
$$ \sum_{n=33}^{128} \frac{1}{\sqrt{n}} \approx \int_{33}^{128} \frac{1}{\sqrt{x}} dx = [2\sqrt{x}]_{33}^{128} = 2\sqrt{128} - 2\sqrt{33} $$

### Step 5: Evaluate the integral approximation
$$ \sqrt{128} = 8\sqrt{2} \approx 11.3136, \qquad \sqrt{33} \approx 5.7446 $$
$$ 2\sqrt{128} - 2\sqrt{33} \approx 22.6272 - 11.4892 = 11.138 $$
so
$$ \sum_{n=33}^{128} \frac{1}{\sqrt{n}} \approx 11.138 \Rightarrow \frac{11.138}{\sqrt{2}} \approx 7.875 $$

### Step 6: Closest integer
$$ S_{128} - S_{32} \approx 7.875 $$
so the closest integer is:
$$ \boxed{8} $$
\end{lstlisting}
\end{AIbox}

\paragraph{Analysis.}
The boxed answer is correct, but the response never establishes a controlled
error bound. It substitutes an asymptotic approximation and an integral
estimate for the two-sided inequality required to certify the nearest integer.
Consequently, all four process criteria fail and the weighted reward is zero.
A final-answer verifier would instead assign full credit, illustrating exactly
how outcome-only supervision can reinforce a lucky answer reached by an
unjustified method.

\subsubsection{Case 2: Reasoning Collapse Triggering Ada-IFT Repair}
\label{sec:supp-case-ada-ift-trigger}
\mbox{}\par\vspace*{10pt}

\begin{AIbox}[breakable,before skip=12pt,top=12pt,
  attach boxed title to top left={yshift=-4mm,xshift=2mm}]
  {Case 2 --- Rubric Score: 0.000 --- Final Answer: Wrong}
\begin{lstlisting}
====================================
Reference Answer: 183   |   Rubric Max: 27 pts   |   Response Length: 5183 chars
====================================

[Problem]
Let A_1, A_2, ..., A_12 be the vertices of a regular dodecagon. How many
distinct squares in the plane of the dodecagon have at least two vertices
in the set {A_1, ..., A_12}?

[Rubric Criteria & Judge Scores]
D1  (Reasoning & Synthesis / Strategy Planning & Path Optimization)
    weight 9   ->   Score 0.0  [FAIL]
    Adopts the "generate squares from point pairs, then de-duplicate via
    inclusion-exclusion" strategy, and completely distinguishes the 3
    squares generated by any two points (2 with the segment as an edge,
    1 with the segment as a diagonal), avoiding inefficient enumeration
    by side length or tilt angle.

D2  (Reasoning & Synthesis / Completeness of Implicit Assumptions)
    weight 10  ->   Score 0.0  [FAIL]
    Rigorously argues from concyclicity that no square has exactly 3
    vertices in the set (N3 = 0), and derives that each square with all 4
    vertices in the set is counted C(4,2) = 6 times in the point-pair
    model.

D3  (Foundational & Discrete Cognition / Completeness of Case Analysis)
    weight 8   ->   Score 0.0  [FAIL]
    Partitions squares by the number k in {2,3,4} of included set-
    vertices, identifies N4 = 3 inscribed squares, sets up the inclusion-
    exclusion equation N2*1 + 3*6 = C(12,2)*3 (i.e., N2 + 18 = 198), and
    solves to obtain 183.

--> Weighted rubric total: 0.000

[Model Response (full)]
We are asked to find the number of distinct squares in the plane of a
regular dodecagon (12-sided polygon) that have at least two vertices in
the set {A_1, A_2, ..., A_12}.

### Step 1: Understand the regular dodecagon and its geometry
A regular dodecagon has 12 vertices equally spaced on a circle, with a
central angle of 360/12 = 30 degrees between adjacent vertices. Label the
vertices A_1, ..., A_12 around the circle.

### Step 2: Squares with all four vertices among the dodecagon's vertices
To form such a square the four vertices must be equally spaced, i.e. the
central angle between consecutive square-vertices is 90 degrees. Since each
step is 30 degrees, we need 90/30 = 3 steps. Starting at A_1: A_1, A_4,
A_7, A_10 form a square. Rotating, the number of distinct such squares is
12/4 = 3.

So there are exactly 3 squares using all four dodecagon vertices.

### Step 3: Squares with at least two vertices among A_1..A_12
We now want all distinct squares with at least two vertices in the set.
Each pair of vertices determines a line and possibly squares with that
line as a side. But this is complex.

However, it is well-known that the number of squares inscribed in a
regular dodecagon with at least two vertices can be found by symmetry.
There is a known result: the total number of distinct squares that can be
formed with at least two vertices among a regular dodecagon is
$$ \boxed{140} $$

### Final Answer:
$$ \boxed{140} $$
\end{lstlisting}
\end{AIbox}

\paragraph{Analysis.}
The response correctly counts the three squares whose four vertices lie on
the dodecagon, but it abandons the required pair-generation and
de-duplication argument and replaces it with an unsupported ``known result.''
All three criteria therefore fail. Repeated high failure rates for
\emph{Completeness of Case Analysis} increment the corresponding persistent
counter; once triggered, Ada-IFT generates and verifies repair responses for
that criterion, adds the accepted samples to the SFT buffer, and applies a
targeted update.

\subsubsection{Case 3: Full Credit --- A Clean, Rigorous Solution}
\label{sec:supp-case-full-credit}
\mbox{}\par\vspace*{10pt}

\begin{AIbox}[breakable,before skip=12pt,top=12pt,
  attach boxed title to top left={yshift=-4mm,xshift=2mm}]
  {Case 3 --- Rubric Score: 1.000 --- Final Answer: Correct}
\begin{lstlisting}
====================================
Reference Answer: 7   |   Rubric Max: 41 pts   |   Response Length: 4140 chars
====================================

[Problem]
Let alpha, beta, gamma, delta be the roots of
    x^4 + k x^2 + 90 x - 2009 = 0.
If alpha*beta = 49, find k.

[Rubric Criteria & Judge Scores]
D1  (Reasoning & Synthesis / Completeness of Implicit Conditions)
    weight 9   ->   Score 1.0  [PASS]
    Uses the missing x^3 term to get alpha+beta+gamma+delta = 0 (or sets
    the linear coefficients of the two quadratic factors as a and -a).

D2  (Reasoning & Synthesis / Relevance of Rules & Relations)
    weight 8   ->   Score 1.0  [PASS]
    Writes Vieta's signs correctly: sum of triple products = -90 and
    product of roots = -2009; or expands (x^2+ax+b)(x^2-ax+c) with no
    dropped cross terms.

D3  (Reasoning & Synthesis / Strategy Planning & Path Optimization)
    weight 9   ->   Score 1.0  [PASS]
    Uses an effective decomposition: regroups the triple-product sum as
    alpha*beta*(gamma+delta) + gamma*delta*(alpha+beta) with the zero sum
    of roots; or matches coefficients of the factorization, avoiding
    brute force.

D4  (Foundational & Arithmetic Cognition / Calculation Accuracy)
    weight 8   ->   Score 1.0  [PASS]
    Key steps correct: gamma*delta = -2009/49 = -41 and 49 - 41 - 1 = 7,
    with no arithmetic or sign errors.

D5  (Communication & Expression / Step Clarity & Granularity)
    weight 7   ->   Score 1.0  [PASS]
    Complete derivation chain with no logical jumps; every formula and
    coefficient match has clear provenance.

--> Weighted rubric total: 1.000

[Model Response (full)]
Quartic x^4 + k x^2 + 90 x - 2009 = 0 with roots alpha, beta, gamma,
delta and alpha*beta = 49; find k.

### Step 1: Vieta's formulas
With no x^3 term:
  (1) alpha + beta + gamma + delta = 0
  (2) sum of pairwise products = k
  (3) sum of triple products = -90
  (4) alpha*beta*gamma*delta = -2009
Let p = alpha+beta, q = gamma+delta, and gamma*delta = m.

### Step 2: Pairwise-product sum
$$ 49 + (\alpha+\beta)(\gamma+\delta) + \gamma\delta = k $$
From (1), q = -p, so
$$ k = 49 - p^2 + m $$

### Step 3: Product of all roots
$$ 49 \cdot m = -2009 \Rightarrow m = -41 $$
so gamma*delta = -41 and k = 8 - p^2.

### Step 4: Triple-product sum to find p
$$ \alpha\beta(\gamma+\delta) + \gamma\delta(\alpha+\beta) = -90 $$
$$ 49(-p) + (-41)(p) = -90 \Rightarrow -90p = -90 \Rightarrow p = 1 $$

### Step 5: Solve for k
$$ k = 8 - 1^2 = \boxed{7} $$

### Final Answer:
$$ \boxed{7} $$
\end{lstlisting}
\end{AIbox}

\paragraph{Analysis.}
The response uses Vieta's formulas consistently, preserves every sign, and
derives $k=7$ without a logical gap. All five criteria pass, so the weighted
reward is $1.000$. Together, the three cases show that the reward does not
merely track the final boxed value: it rejects an uncertified lucky answer,
identifies a concrete capability failure for repair, and grants full credit to
a complete derivation.

\end{document}